\documentclass{bmcart}

\usepackage{graphicx}
\usepackage{amsmath,amssymb,amsfonts}
\usepackage{algpseudocode,algorithm}
\usepackage{subfigure}

\startlocaldefs
\endlocaldefs

\begin{document}

\begin{frontmatter}

\begin{fmbox}
\dochead{Research}


\title{Optimization Algorithm for Feedback and Feedforward Policies towards Robot Control Robust to Sensing Failures}


\author[
   addressref={naist},                   
   corref={naist},                       
   email={kobayashi@is.naist.jp}   
]{\inits{TK}\fnm{Taisuke} \snm{Kobayashi}}
\author[
   addressref={naist},
]{\inits{KY}\fnm{Kenta} \snm{Yoshizawa}}


\address[id=naist]{
  \orgname{Division of Information Science, Graduate School of Science and Technology, Nara Institute of Science and Technology}, 
  \street{Takayama-cho},                     %
  \postcode{630-0192}                                
  \city{Nara},                              
  \cny{Japan}                                    
}


\begin{artnotes}
\end{artnotes}

\end{fmbox}


\begin{abstractbox}

\begin{abstract} 
\parttitle{Background and problem statement} 
Model-free or learning-based control, in particular, reinforcement learning (RL), is expected to be applied for complex robotic tasks.
Traditional RL requires a policy to be optimized is state-dependent, that means, the policy is a kind of feedback (FB) controllers.
Due to the necessity of correct state observation in such a FB controller, it is sensitive to sensing failures.
To alleviate this drawback of the FB controllers, feedback error learning integrates one of them with a feedforward (FF) controller.
RL can be improved by dealing with the FB/FF policies, but to the best of our knowledge, a methodology for learning them in a unified manner has not been developed.

\parttitle{Contribution} 
In this paper, we propose a new optimization problem for optimizing both the FB/FF policies simultaneously.
Inspired by \textit{control as inference}, the optimization problem considers minimization/maximization of divergences between trajectory, predicted by the composed policy and a stochastic dynamics model, and optimal/non-optimal trajectories.
By approximating the stochastic dynamics model using variational method, we naturally derive a regularization between the FB/FF policies.
In numerical simulations and a robot experiment, we verified that the proposed method can stably optimize the composed policy even with the different learning law from the traditional RL.
In addition, we demonstrated that the FF policy is robust to the sensing failures and can hold the optimal motion.

\end{abstract}




\end{abstractbox}
%

\end{frontmatter}

\section{Introduction}
\label{sec:introduction}

In the last decade, the tasks (or objects) required of robots have become steadily more complex.
For such next-generation robot control problems, traditional model-based control like~\cite{kobayashi2018unified} seems to reach its limit due to the difficulty of modeling complex systems.
Model-free or learning-based control like~\cite{itadera2021towards} is expected to resolve these problems in recent year.
In particular, reinforcement learning (RL)~\cite{sutton2018reinforcement} is one of the most promising approaches to this end, and indeed, RL integrated with deep neural networks~\cite{lecun2015deep}, so-called deep RL~\cite{mnih2015human}, achieved several complex tasks:
e.g. human-robot interaction~\cite{modares2015optimized}; manipulation of deformable objects~\cite{tsurumine2019deep}; and manipulation of various general objects from scratch~\cite{kalashnikov2018scalable}.

In principle, RL makes an agent to optimize a policy (a.k.a. controller) to stochastically sample action (a.k.a. control input) depending on state, result of interaction between the agent and environment~\cite{sutton2018reinforcement}.
Generally speaking, therefore, the policy to be optimized can be regarded as one of the feedback (FB) controllers.
Of course, the policy is more conceptual and general than traditional FB controllers such as for regulation and tracking, but it is still a mapping from state to action.

Such a FB policy inherits the drawbacks of the traditional FB controllers, i.e. the sensitivity to sensing failures~\cite{sugimoto2020relaxation}.
For example, if the robot has a camera to detect an object, pose of which is given to be state of RL, the FB policy would sample erroneous action according to a wrong pose by occlusion.
Alternatively, if the robot system is connected with a wireless TCP/IP network to sense data from IoT devices, communication loss or delay due to poor signal conditions will occur at irregular intervals, causing erroneous action.

To alleviate this fundamental problem of the FB policy, previous studies have developed the policies that do not depend only on state.
In a straightforward way, time-dependent policy has been proposed by directly adding the elapsed time to state~\cite{musial2007feed} or by utilizing recurrent neural networks (RNNs)~\cite{hochreiter1997long,murata2013learning} for approximation of that policy~\cite{lee2020stochastic}.
If the policy is computed according to the phase and spectrum information of the system, instantaneous sensing failures can be ignored~\cite{sharma2018phase,azizzadenesheli2016reinforcement}.
In an extreme case, if the robot learns to episodically generate the trajectory, the adaptive behavior to state is completely lost, but it is never affected by the sensing failures.

From the perspective of the traditional control theory and biology, it has been suggested that this problem of the FB policy can be resolved by a feedforward (FF) policy with feedback error learning (FEL)~\cite{miyamoto1988feedback,nakanishi2004feedback,sugimoto2008feedback,sugimoto2020relaxation}.
FEL is a framework in which the FF controller is updated based on the error signal of the FB controller, and finally the control objective is achieved only by the FF controller.
In other words, instead of designing only the single policy as in the previous studies above, FEL has both the FB/FF policies in the system and composes their outputs appropriately to complement each other's shortcomings: the sensitivity to the sensing failures in the FB policy; and the lack of adaptability to the change of state in the FF policy.
The two separate policies are more compact than the integrated one.
In addition, although the composition of the outputs in the previous studies is a simple summation, it creates a new room for designing different composition rules, which makes it easier for designers to adjust which of the FB/FF policies is preferred.

The purpose of this study is to  take over the benefits of FEL to the RL framework, as shown in Fig.~\ref{fig:fffb_framework}.
To this end, we have to solve two challenges as below.
\begin{enumerate}
    \item Since RL is not only for tracking problem, which is the target of FEL, we need to design how to compose the FB/FF policies.
    \item Since the FB policy is not fixed unlike FEL, both of the FB/FF policies are required to be optimized simultaneously.
\end{enumerate}

For the first challenge, we assumes that the composed policy is designed as mixture distribution of the FB/FF policies since RL policy is stochastically defined.
In addition, we heuristically design its mixture ratio depending on confidences of the respective FB/FF policies so that the higher confident policy is prioritized.

For the second challenge, inspired by \textit{control as inference}~\cite{levine2018reinforcement}, we derive a new optimization problem to minimize/maximize the divergences between trajectory, predicted by the composed policy and a stochastic dynamics model, and optimal/non-optimal trajectories.
Furthermore, by designing the stochastic dynamics model with variational approximation~\cite{chung2015recurrent}, we yield regularization between the FB/FF policies.
We expect that skill of the FB policy, which can be optimized faster than the FF policy, will be transferred into the FF policy via this regularization.

To verify that the proposed method can optimize the FB/FF policies in a unified manner, we conduct numerical simulations for statistical evaluation and a robot experiment as demonstration.
Through the numerical simulations, we show the capability of the proposed method, namely, stable optimization of the composed policy even with the different learning law from the traditional RL.
However, the proposed method occasionally fails to learn the optimal policy.
We analyze this reason as the extreme updating of the FF policy (or RNNs) to wrong direction.
In addition, after training on the robot experiment, we clarify the value of the proposed method that the optimized FF policy robustly samples valuable actions to the sensing failures even when the FB policy fails to achieve the optimal behavior.

\section{Preliminaries}

\subsection{Reinforcement learning}

In RL~\cite{sutton2018reinforcement}, an agent interacts with unknown environment using action $a \in \mathcal{A}$ sampled from policy $\pi$.
The environment returns the result of the interaction as state $s \in \mathcal{S}$ and evaluates it according to reward function $r(s, a) \in \mathbb{R}$.
The optimization problem of RL is to find the optimal policy $\pi^*$ that maximizes the sum of rewards in the future from the current time $t$ (or, called return), defined as $R_t = \sum_{k=0}^\infty \gamma^k r_{t+k}$ with $\gamma \in [0, 1)$ discount factor.

RL generally assumes that the environment follows Markov process, i.e. the next state $s^\prime$ is sampled from $s^\prime \sim p_e(s^\prime \mid s, a)$.
By additionally limiting the policy as $\pi(a \mid s)$, Markov decision process (MDP) is satisfied.
In that case, RL can be illustrated as the agent-environment-loop at the top of Fig.~\ref{fig:problem_rl}.
However, in practical use, measurement of state causes delay (e.g. due to overload in the communication networks) and/or loss (e.g. occlusion in camera sensors), suggested in the bottom of Fig.~\ref{fig:problem_rl}.
To solve this problem, this paper therefore proposes a new method to optimize the FB/FF policies in a unified manner by formulating them without necessarily requiring MDP.

In the conventional RL under MDP, the expected value of $R$ is functionalized as $V(s)$ as (state) value function and $Q(s, a)$ as (state-)action value function, and $V$ can be learned by the following equation.
\begin{align}
    \delta &= Q(s,a) - V(s) \simeq r(s, a) + \gamma V(s^\prime) - V(s)
    \label{eq:td_err} \\
    \mathcal{L}_\mathrm{value} &= \cfrac{1}{2}\delta^2
    \label{eq:loss_value}
\end{align}
Note that $Q$ can also be learned with the similar equation, although we do not use $Q$ directly in this paper.

Based on $\delta$, an actor-critic algorithm~\cite{konda2000actor} updates $\pi$ according to the following policy gradient.
\begin{align}
    \nabla \mathcal{L}_\pi &= - \mathbb{E}_{p_e \pi} [\delta \nabla \ln \pi(a \mid s)]
    \label{eq:grad_ac}
\end{align}
where $\mathbb{E}_{p_e \pi} [\cdot]$ is approximated by Monte Carlo method.

\subsection{Introduction of optimality variable in \textit{control as inference}}

Recently, RL can be regarded as inference problem, so-called control as inference~\cite{levine2018reinforcement}.
This extension of interpretation is realized by introducing a optimality variable, $o = \{0, 1\}$, which represents whether the current state $s$ and action $a$ are optimal ($o = 1$) or not ($o = 0$).
Since it is defined as random variable, the probability of $o = 1$, $p(o=1 \mid s, a)$, is parameterized by reward $r$ to connect the conventional RL with this interpretation.
\begin{align}
    p(o = 1 \mid s, a) = \exp\left( \cfrac{r(s, a) - c}{\tau} \right)
    \label{eq:def_opt_r}
\end{align}
where $c = \max(r)$ to satisfy $e^{r(s, a) - c} \leq 1$, and $\tau$ denotes the hyperparameter to clarify uncertainty, and can be adaptively tuned.

Furthermore, by considering the optimality in the future as $O$, we can connect this formulation with the conventional value functions.
Specifically, the following probability can be derived.
\begin{align}
    p(O = 1 \mid s) &= \exp\left( \cfrac{V(s) - C}{\tau} \right)
    \label{eq:def_opt_V} \\
    p(O = 1 \mid s, a) &= \exp\left( \cfrac{Q(s, a) - C}{\tau} \right)
    \label{eq:def_opt_Q}
\end{align}
where $C = \max(V) = \max(Q)$ theoretically, although its specific value is generally unknown.

In this way, the optimality can be treated in probabilistic inference problems, facilitating integration with such as Bayesian inference and other methods.
This paper utilizes this property to derive a new optimization problem, as derived later.

\subsection{Variational recurrent neural network}

To reveal state transition probability (i.e. $p_e$) as stochastic dynamics model, we derive the method to learn it based on variational recurrent neural network (VRNN)~\cite{chung2015recurrent}.
Therefore, in this section, we briefly introduce the VRNN.

The VRNN considers the maximization problem of log likelihood of a prediction model of observation ($s$ in the context of RL), $p_m$.
$s$ is assumed to be stochastically decoded from lower-dimensional latent variable $z$, and $z$ is also sampled according to the history of $s$, $h^s$, as time-dependent prior $p(z \mid h^s)$.
Here, $h^s$ is generally approximated by recurrent neural networks, and this paper employs deep echo state networks~\cite{gallicchio2018design} for this purpose.
Using Jensen's inequality, a variational lower bound is derived as follows:
\begin{align}
    \ln p_m(s \mid h^s) &= \ln \int p(s \mid z) p(z \mid h^s) dz
    \nonumber \\
    &= \ln \int q(z \mid s, h^s) p(s \mid z) \cfrac{p(z \mid h^s)}{q(z \mid s, h^s)} dz
    \nonumber \\
    &\geq \mathbb{E}_{q(z \mid s, h^s)}[\ln p(s \mid z)]
    \nonumber \\
    &- \mathrm{KL}(q(z \mid s, h^s) \| p(z \mid h^s))
    \nonumber \\
    &= - \mathcal{L}_\mathrm{vrnn}
    \label{eq:loss_vrnn}
\end{align}
where $p(s \mid z)$ and $q(z \mid s, h^s)$ denote the decoder and encoder, respectively.
$\mathrm{KL}(\cdot \| \cdot)$ is the term for Kullback-Leibler (KL) divergence between two probabilities.
$\mathcal{L}_\mathrm{vrnn}$ is minimized via the optimization of $p_m$, which consists of $p(s \mid z)$, $q(z \mid s, h^s)$, and $p(z \mid h^s)$.

Note that, in the original implementation\cite{chung2015recurrent}, the decoder is also depending on $h^s$, but that is omitted in the above derivation for simplicity and for aggregating time information to $z$.
In addition, the strength of regularization by the KL term can be controlled by following $\beta$-VAE~\cite{higgins2017beta} with a hyperparameter $\beta \geq 0$.

\section{Derivation of proposed method}

\subsection{Overview}

The outputs of FB/FF policies should eventually coincide, but it is unclear how they will be updated if we directly optimize the composed policy according to the conventional RL.
In this paper, we propose a unified optimization problem in which the FB/FF policies naturally coincide and the composed one is properly optimized.
The key points in the proposed method are two folds:
\begin{enumerate}
    \item The trajectory predicted with the stochastic dynamics model and the composed policy is expected to be close to/away from optimal/non-optimal trajectories inferred with the optimality variable.
    \item The stochastic dynamics model is trained via its variational lower bound, which naturally generates a soft constraint between the FB/FF policies.
\end{enumerate}

Here, as an additional preliminary preparation, we define the FB, FF, and composed policies mathematically: $\pi_\mathrm{FB}(a \mid s)$; $\pi_\mathrm{FF}(a \mid h^a)$; and the following mixture distribution, respectively.
\begin{align}
    \pi(a \mid s, h^a) = w \pi_\mathrm{FB}(a \mid s) + (1 - w) \pi_\mathrm{FF}(a \mid h^a)
    \label{eq:def_policy_mix}
\end{align}
where $w \in [0, 1]$ denotes the mixture ratio of the FB/FF policies.
That is, for generality, the outputs of the FB/FF policies are composed by a stochastic switching mechanism, rather than a simple summation as in FEL~\cite{miyamoto1988feedback}.
Note that since the history of action, $h^a$, can be updated without $s$, the FF policy is naturally robust to sensing failures.

\subsection{Inference of optimal/non-optimal policies}

First of all, we infer the optimal policy, which yields the optimal trajectory by interacting with the real environment $p_e$, and the non-optimal policy, which causes the non-optimal trajectory on the contrary.
With eqs.~\eqref{eq:def_opt_V} and~\eqref{eq:def_opt_Q}, the policy conditioned on $O$, $\pi^*(a \mid s, h^a, O)$, can be derived through Bayes theorem.
\begin{align}
    \pi^*(a \mid s, h^a, O) = \cfrac{p(O \mid s, a) b(a \mid s, h^a)}{p(O \mid s)}
    \label{eq:policy_cond}
\end{align}
where $b(a \mid s, h^a)$ denotes the sampler distribution (e.g. the composed policy with old parameters or one approximated by target networks~\cite{kobayashi2021t}).

By substituting $\{0,1\}$ for $O$, the inference of the optimal policy, $\pi^+$, and the non-optimal policy, $\pi^-$ is given as follows:
\begin{align}
    \pi^+(a \mid s, h^a) &= \pi^*(a \mid s, h^a, O=1)
    = \cfrac{\exp\left( \cfrac{Q(s,a) - C}{\tau} \right)}{\exp\left( \cfrac{V(s) - C}{\tau} \right)} b(a \mid s, h^a)
    \label{eq:def_policy_opt} \\
    \pi^-(a \mid s, h^a) &= \pi^*(a \mid s, h^a, O=0)
    = \cfrac{1 - \exp\left( \cfrac{Q(s,a) - C}{\tau} \right)}{1 - \exp\left( \cfrac{V(s) - C}{\tau} \right)} b(a \mid s, h^a)
    \label{eq:def_policy_nopt}
\end{align}
Although it is difficult to sample action from these policies directly, they can be utilized for analysis in the next section.

\subsection{Optimization problem for optimal/non-optimal trajectories}

With the composed policy, $\pi$, and the stochastic dynamics model, given as $p_m(s^\prime \mid s, a, h^s, h^a)$, a part of trajectory is predicted as $p_m \pi$.
As a reference, we can consider the part of trajectory with $\pi^*$ in eq.~\eqref{eq:policy_cond} and the real environment, $p_e$, as $p_e \pi^*$.
The degree of divergence between the two can be evaluated by KL divergence as follows:
\begin{align}
    \mathrm{KL}(p_e \pi^* \| p_m \pi) &= \mathbb{E}_{p_e \pi^*} [(\ln p_e + \ln \pi^*) - (\ln p_m + \ln \pi)]
    \nonumber \\
    &= \mathbb{E}_{p_e b} \left [ \cfrac{p(O \mid s, a)}{p(O \mid s)} \{(\ln p_e + \ln \pi^*) - (\ln p_m + \ln \pi)\} \right ]
    \nonumber \\
    &\propto - \mathbb{E}_{p_e b} \left [ \cfrac{p(O \mid s, a)}{p(O \mid s)} (\ln p_m + \ln \pi) \right ]
    \label{eq:kl_traj_cond}
\end{align}
where the term $\ln p_e \pi^*$ inside the expectation operation is excluded since it is not related to the learnable $p_m$ and $\pi$.
The expectation operation with $p_e$ and $b$ can be approximated by Monte Carlo method, namely, we can optimize $p_m$ and $\pi$ using the above KL divergence with the appropriate conditions of $O$.

As the conditions, our optimization problem considers that $p_m \pi$ is expected to be close to $p_e \pi^+$ (i.e. the optimal trajectory) and be away from $p_e \pi^-$ (i.e. the non-optimal trajectory), as shown in Fig.~\ref{fig:problem_traj_div}.
Therefore, the specific loss function to be minimized is given as follows:
\begin{align}
    \mathcal{L}_\mathrm{traj} &= \mathrm{KL}(p_e \pi^+ \mid p_m \pi) - \mathrm{KL}(p_e \pi^- \mid p_m \pi)
    \nonumber \\
    &\propto - \mathbb{E}_{p_e b}\left [ \left \{ \cfrac{\exp\left( \cfrac{Q - C}{\tau} \right)}{\exp\left( \cfrac{V - C}{\tau} \right)} - \cfrac{1 - \exp\left( \cfrac{Q - C}{\tau} \right)}{1 - \exp\left( \cfrac{V - C}{\tau} \right)} \right \} (\ln p_m + \ln \pi) \right ]
    \nonumber \\
    &= - \mathbb{E}_{p_e b}\left [ \cfrac{\exp\left( \cfrac{Q - V}{\tau} \right) - 1}{1 - \exp \left( \cfrac{V - C}{\tau} \right)} (\ln p_m + \ln \pi) \right ]
    \nonumber \\
    &\propto - \mathbb{E}_{p_e b}\left [ \tau \left \{\exp\left( \cfrac{\delta}{\tau} \right) - 1 \right \} (\ln p_m + \ln \pi) \right ]
    \label{eq:loss_traj}
\end{align}
where $1 - \exp \{(V - C)\tau^{-1}\}$ and $\tau$ are multiplied to eliminate unknown $C$ and to scale the gradient at $\delta = 0$ to be one, respectively.
Note that the derived result is similar to eq.~\eqref{eq:grad_ac}, but with a different coefficient from $\delta$ and a different sampler from $\pi$.

\subsection{Stochastic dynamics model with variational lower bound}

In eq.~\eqref{eq:loss_traj}, $\ln p_m$, i.e. the stochastic dynamics model, is included and it should be modeled.
Indeed, we found that the model based on the VRNN~\cite{chung2015recurrent} shown in eq.~\eqref{eq:loss_vrnn} can naturally yield an additional regularization between the FB/FF policies.
In addition, such a method is regarded as one for extracting latent Markovian dynamics in problems for which MDP is not established in the observed state, and is similar to the latest model-based RL~\cite{chua2018deep,clavera2019model}.

Specifically, we consider the dynamics of latent variable $z$ as $z^\prime = f(z, a)$ with $f$ learnable function, and $a$ can be sampled from time-dependent prior (i.e. the FF policy).
In that time, eq.~\eqref{eq:loss_vrnn} is modified through the following derivation.
\begin{align}
    \ln p_m(s^\prime \mid h^s, h^a) &= \ln \iint p(s^\prime \mid z^\prime) p(z \mid h^s) \pi_\mathrm{FF}(a \mid h^a) dz da
    \nonumber \\
    &= \ln \iint q(z \mid s, h^s) \pi(a \mid s, h^a) p(s^\prime \mid z^\prime)
    \nonumber \\
    &\times \cfrac{p(z \mid h^s)}{q(z \mid s, h^s)} \cfrac{\pi_\mathrm{FF}(a \mid h^a)}{\pi(a \mid s, h^a)} dz da
    \nonumber \\
    &\geq \mathbb{E}_{q(z \mid s, h^s) \pi(a \mid s, h^a)} [ \ln p(s^\prime \mid z^\prime) ]
    \nonumber \\
    &- \mathrm{KL}(q(z \mid s, h^s) \| p(z \mid h^s))
    - \mathrm{KL}(\pi(a \mid s, h^a) \| \pi_\mathrm{FF}(a \mid h^a))
    \nonumber \\
    &= - \mathcal{L}_\mathrm{model}
    \label{eq:loss_model}
\end{align}
Since we know the composed policy $\pi$ is mixture of the FB/FF policies defined in eq.~\eqref{eq:def_policy_mix}, the KL term between $\pi$ and $\pi_\mathrm{FF}$ can be decomposed using variational approximation~\cite{hershey2007approximating} and Jensen's inequality.
\begin{align}
    \mathrm{KL}(\pi \| \pi_\mathrm{FF}) &\geq w \ln \cfrac{w e^{-\mathrm{KL}(\pi_\mathrm{FF} \| \pi_\mathrm{FF})}
    + (1 - w) e^{-\mathrm{KL}(\pi_\mathrm{FB} \| \pi_\mathrm{FF})}}
    {e^{-\mathrm{KL}(\pi_\mathrm{FB} \| \pi_\mathrm{FF})}}
    \nonumber \\
    &+ (1 - w) \ln \cfrac{w e^{-\mathrm{KL}(\pi_\mathrm{FF} \| \pi_\mathrm{FB})}
    + (1 - w) e^{-\mathrm{KL}(\pi_\mathrm{FF} \| \pi_\mathrm{FF})}}
    {e^{-\mathrm{KL}(\pi_\mathrm{FF} \| \pi_\mathrm{FF})}}
    \nonumber \\
    &= w \ln \{ w e^{\mathrm{KL}(\pi_\mathrm{FB} \| \pi_\mathrm{FF})} + (1 - w) \}
    \nonumber \\
    &+ (1 - w) \ln \{ w e^{-\mathrm{KL}(\pi_\mathrm{FF} \| \pi_\mathrm{FB})} + (1 - w) \}
    \nonumber \\
    &\geq w^2 \mathrm{KL}(\pi_\mathrm{FB} \| \pi_\mathrm{FF})
    - (1 - w) w \mathrm{KL}(\pi_\mathrm{FF} \| \pi_\mathrm{FB})
    \nonumber \\
    &= w^2 \{H(\pi_\mathrm{FB} \| \pi_\mathrm{FF}) - H(\pi_\mathrm{FB})\}
    - (1 - w) w \mathrm{KL}(\pi_\mathrm{FF} \| \pi_\mathrm{FB})
    \nonumber \\
    &\propto w^2 H(\pi_\mathrm{FB} \| \pi_\mathrm{FF})
\end{align}
where we use the fact that $\mathrm{KL}(p \| q) = H(p \| q) - H(p)$ with $H(\cdot \| \cdot)$ cross entropy and $H(\cdot)$ (differential) entropy.
By eliminating the negative KL term and the negative entropy term, which are unnecessary for regularization, only the cross entropy remains.

The general case of VAE omits the expectation operation by sampling only one $z$ (and $a$ in the above case) according to $s$.
In addition, as explained before, the strength of regularization can be controlled by adding $\beta$~\cite{higgins2017beta}.
With this fact, we can modify $\mathcal{L}_\mathrm{model}$ as follows:
\begin{align}
    \mathcal{L}_\mathrm{model} = - \ln p(s^\prime \mid z^\prime)
    + \beta_z \mathrm{KL}(q(z \mid s, h^s) \| p(z \mid h^s))
    + \beta_a w^2 H(\pi_\mathrm{FB} \| \pi_\mathrm{FF})
    \label{eq:loss_model2}
\end{align}
where $z \sim q(z \mid s, h^s), a \sim \pi(a \mid s, h^a), z^\prime = f(z, a)$, and $\beta_{z,a}$ denote the strength of regularization for each.
Finally, the above $\mathcal{L}_\mathrm{model}$ can be substituted into eq.~\eqref{eq:loss_traj} as $- \ln p_m$.
\begin{align}
    \mathcal{L}_\mathrm{traj} = - \mathbb{E}_{p_e b}\left [ \tau \left \{\exp\left( \cfrac{\delta}{\tau} \right) - 1 \right \} (- \mathcal{L}_\mathrm{model} + \ln \pi) \right ]
    \label{eq:loss_traj2}
\end{align}

As can be seen in eq.~\eqref{eq:loss_model2}, the regularization between the FB/FF policies is naturally added.
Its strength is depending on $w^2$, that is, as the FB policy is prioritized (i.e. $w$ is increased), this regularization is reinforced.
In addition, since $\mathcal{L}_\mathrm{model}$ is now inside of $\mathcal{L}_\mathrm{traj}$, the regularization becomes strong only when $\delta > 0$ enough, that is, the agent knows the optimal direction for updating $\pi$.
Usually, at the beginning of RL, the policy generates random actions, which make the FF policy be optimized;
in contrast, the FB policy can be optimized under weak regularization (if the observation is sufficiently performed).
Afterwards, if $w$ is adaptively given (as introduced in the next section), the FB policy will be strongly connected with the FF policy.
In summary, with this formulation, we can expect that the FB policy will be optimized first while regularization is weakened, and that its skill will gradually be transferred to the FF policy as like FEL~\cite{miyamoto1988feedback}.

\section{Additional design for implementation}

\subsection{Design of mixture ratio based on policy entropy}

For the practical implementation, we first design the mixture ratio $w \in [0, 1]$ heuristically.
As its requirements, the composed policy should prioritize the policy with higher confidence from the FB/FF policies.
In addition, if the FB/FF policies are similar to each other, either can be selected.
Finally, even for arbitrary distribution model of the FB/FF policies, $w$ must be computable.

As one of the solutions for these requirements, we design the following $w$ with the entropies for the FB/FF policies, $H_\mathrm{FB}, H_\mathrm{FF}$, and the L2 norm between the means of these policies, $d = \| \mu_\mathrm{FB} - \mu_\mathrm{FF} \|_2$.
\begin{align}
    w = \cfrac{\exp(-H_\mathrm{FB} d \beta_T)}{\exp(-H_\mathrm{FB} d \beta_T) + \exp(-H_\mathrm{FF} d \beta_T)}
    \label{eq:policy_mix_ratio}
\end{align}
where $\beta_T > 0$ denotes the inverse temperature parameter, i.e. $w$ tends to be deterministic at $0$ or $1$ with higher $\beta_T$; and vice versa.
Note that as lower entropy has higher confidence, the negative entropies are applied into softmax function.

If one of the entropies is sufficiently smaller than another, $w$ will converge on $1$ or $0$ for prioritizing the FB/FF policies, respectively.
However, if these policies output similar values on average, the robot can select action from either policy, so the inverse temperature is adaptively lowered by $d$ to make $w$ converge to $w \simeq 0.5$.

\subsection{Partial cut of computational graph}

In general, VAE-based architecture holds the computational graph, which gives paths for backpropagation, of latent variable $z$ by reparameterization trick.
If this trick is applied to $a$ in our dynamics model as it is, the policy $\pi$ will be updated toward one for improving the prediction accuracy, not for maximizing the return, which is the original purpose of policy optimization in RL.

To mitigate the wrong updates of $\pi$ while preserving the capability to backpropagate the gradients to the whole network as in VAE, we partially cut the computational graph as follows:
\begin{align}
    a \gets \eta a + (1 - \eta) \hat{a}
    \label{eq:action_graph_cut}
\end{align}
where $\eta$ denotes the hyperparameter and $\hat{\cdot}$ cuts the computational graph and represents merely value.

\subsection{Auxiliary loss functions}

As can be seen in eq.~\eqref{eq:loss_traj2}, if $\delta < 0$, $- \mathcal{L}_\mathrm{model}$ will be minimized, reducing the prediction accuracy of dynamics.
As for the policy, it is desirable to have a sign reversal of its loss according to $\delta$ to determine whether the update direction is good or bad.
On the other hand, since the dynamics model should ideally have a high prediction accuracy for any state, this update rule may cause the failure of optimization.

In order not to reduce the prediction accuracy, we add an auxiliary loss function.
We focus on the fact that the lower bound of the coefficient in eq.~\eqref{eq:loss_traj2}, $\tau (\exp(\delta\tau^{-1}) - 1)$, is bounded and can be found analytically to be $-\tau$ when $\delta \to - \infty$.
That is, by adding $\tau \mathcal{L}_\mathrm{model}$ as the auxiliary loss function, the dynamics model should be updated toward one with higher prediction accuracy, while its update amount is still weighted by $\exp(\delta\tau^{-1})$.

To update the value function, $V$, the conventional RL uses eq.~\eqref{eq:loss_value}.
Instead of it, we found that the minimization problem of the KL divergence between $p(O \mid s, a)$ and $p(O \mid s)$ yields the following loss function similar to eq.~\eqref{eq:loss_traj2}.
\begin{align}
    \mathcal{L}_\mathrm{value} = - \mathbb{E}_{p_e b}\left [ \tau \left \{\exp\left( \cfrac{\delta}{\tau} \right) - 1 \right \} V \right ]
    \label{eq:loss_value2}
\end{align}
Note that, in this formula (and eq.~\eqref{eq:loss_traj2}), $\delta$ has no computational graph for backpropagation, i.e. it is merely coefficient.

Finally, the loss function to be minimized for updating $\pi$ (i.e. $\pi_\mathrm{FB}$ and $\pi_\mathrm{FF}$), $V$, and $p_m$ can be summarized as follows:
\begin{align}
    \mathcal{L}_\mathrm{all} = \mathcal{L}_\mathrm{traj} + \mathcal{L}_\mathrm{value} + \tau \mathcal{L}_\mathrm{model}
    \label{eq:loss_all}
\end{align}
where $\mathcal{L}_\mathrm{traj}$, $\mathcal{L}_\mathrm{value}$, and $\mathcal{L}_\mathrm{model}$ are given in eqs.~\eqref{eq:loss_traj2}, \eqref{eq:loss_value2}, and \eqref{eq:loss_model2}, respectively.
This loss function can be minimized by one of the stochastic gradient descent (SGD) methods like~\cite{ziyin2020laprop}.

\section{Experiment}

\subsection{Objective}

We verify the validity of the proposed method derived in this paper.
This verification is done through a numerical simulation of a cart-pole inverted pendulum and an experiment of a snake robot forward locomotion, which is driven by central pattern generators (CPGs)~\cite{cohen1982nature}.

Four specific objectives are listed as below.
\begin{enumerate}
    \item Through the simulation and the robot experiment, we verify that the proposed method can optimize the composed policy, optimization process of which is also revealed.
    \item By comparing the successful and failing cases in the simulation, we clarify an open issue of the proposed method.
    \item We compare two behaviors with the decomposed FB/FF policies to make sure there is little difference between them.
    \item By intentionally causing sensing failures in the robot experiment, we illustrate the sensitivity/robustness of FB/FF policies to it, respectively.
\end{enumerate}

\subsection{Setup of proposed method}

The network architecture for the proposed method is designed using PyTorch~\cite{paszke2017automatic}, as illustrated in Fig.~\ref{fig:fffb_architecture}.
All the modules (i.e. the encoder $q(z \mid s, h^s)$, decoder $p(s^\prime \mid z^\prime)$, time-dependent prior $q(z \mid h^s)$, dynamics $f(z, a)$, value function $V(s)$, and the FB/FF policies $\pi_\mathrm{FB}(a \mid s)$, $\pi_\mathrm{FF}(a \mid h^a)$) are represented by three fully connected layers with 100 neurons for each.
As nonlinear activation functions for them, we apply layer normalization~\cite{ba2016layer} and Swish function~\cite{elfwing2018sigmoid}.
To represent the histories, $h^s$ and $h^a$, as mentioned before, we employ deep echo state networks~\cite{gallicchio2018design} (three layers with 100 neurons for each).
Probability density function outputted from all the stochastic model is given as student-t distribution with reference to~\cite{takahashi2018student,kobayashi2019variational,kobayashi2019student}.

To optimize the above network architecture, a robust SGD, i.e., LaProp~\cite{ziyin2020laprop} with t-momentum~\cite{ilboudo2020robust} and d-AmsGrad~\cite{kobayashi2021towards} (so-called td-AmsProp), is employed with their default parameters except the learning rate.
In addition, optimization of $V$ and $\pi$ can be accelerated by using adaptve eligibility traces~\cite{kobayashi2020adaptive}, and stabilized by using t-soft target network~\cite{kobayashi2021t}.

The parameters for the above implementation, including those unique to the proposed method, are summarized in Table~\ref{tab:parameter}.
Many of these were empirically adjusted based on values from previous studies.
Because of the large number of parameters involved, the influence of these parameters on the behavior of the proposed method is not examined in this paper.
However, it should be remarked that a meta-optimization of them can be easily performed with packages such as Optuna~\cite{akiba2019optuna}, although such a meta-optimization requires a great deal of time.

\subsection{Simulation for statistical evaluation}

For the simulation, we employ Pybullet dynamics engine wrapped by OpenAI Gym~\cite{coumans2016pybullet,brockman2016openai}.
A task (a.k.a. environment), \textit{InvertedPendulumBullet-v0}, where a cart tries to keep a pole standing on it, is tried to be solved.
With different random seeds, 30 trials involving 300 episodes for each are performed.

First of all, we depict the learning curves about the score (a.k.a. the sum of rewards) and the mixture ratio in Fig.~\ref{fig:sim_result}.
Since five of them were obvious failures, for further analysis, we separately depicted \textit{Failure (5)} for the five failures and \textit{Success (25)} for the remaining successful trials.
We can see in the successful trials that the agent could solve this balancing task stably after 150 episodes, even with stochastic actions.
Furthermore, further stabilization and making the composed policy deterministic were accelerated, and in the end, the task was almost certainly accomplished by the proposed method in the successful 25 trials.

Focusing on the mixture ratio, the FB policy was dominant in the early stages of learning, as expected.
Then, as the episodes passed, the FF policy was optimized toward the FB policy, and the mixture ratio gradually approached 0.5.
Finally, it seems to have converged to around 0.7, suggesting that the proposed method is basically dominated by the FB policy under stable observation.

Although all the trials obtained almost the same curves until 50 episodes in both figures, the failure trials suddenly decreased their scores.
In addition, probably due to the failure of optimization of the FF policy, the mixture ratio in the failure trials fixed on almost 1.
It is necessary to clarify the cause of this apparent difference from the successful trials, i.e. the open issue of the proposed method.

To this end, we decompose the mixture ratio into the distance between the FB/FF policies, $d$, and the entropies of the respective policies, $H_\mathrm{FB}$ and $H_\mathrm{FF}$, in Fig.~\ref{fig:sim_analysis}.
Extreme behavior can be observed around 80th episode in $d$ and $H_\mathrm{FF}$.
This suggests that the FF policy (or its base RNNs) was updated extremely wrong direction, and could not be reverted from there.
As a consequence, the FB policy was also constantly regularized to the FF policy, i.e. the wrong direction, causing the failures of the balancing task.
Indeed, $H_\mathrm{FB}$ was gradually increased toward $H_\mathrm{FF}$.
In summary, the proposed method lacks the stabilization of learning of the FF policy (or its base RNNs).
It is however expected to be improved by suppressing the amount of policy updates like the latest RL~\cite{kobayashi2020proximal}, regularization of RNNs~\cite{zaremba2014recurrent}, and/or promoting initialization of the FF policy.

\subsection{Robot experiment}

The following robot experiment is conducted to illustrate the practical value of the proposed method.
Since the statistical properties of the proposed method are verified via the above simulation, we analyze one successful case here.

\subsubsection{Setup of robot and task}

A snake robot used in this experiment is shown in Fig.~\ref{fig:snake_robot}.
This robot has eight Qbmove actuators developed by QbRobotics, which can control the stiffness in hardware level, i.e. variable stiffness actuator (VSA)~\cite{catalano2011vsa}.
As can be seen in the figure, all the actuators are serially connected and on casters to easily drive by snaking locomotion.
On the head of the robot, a AR marker is attached to detect its coordinates using a camera (ZED2 developed by Stereolabs).

To generate the primitive snaking locomotion, we employ CPGs~\cite{cohen1982nature} as mentioned before.
Each CPG follows Cohen's model with sine function as follows:
\begin{align}
    \zeta_i &= \zeta_i + \left \{ u_i^r + \sum_{ij} \alpha(\zeta_j + \zeta_i - u_i^\eta) \right \} dt
    \label{eq:cpg_dyn} \\
    \theta_i &= u_i^A \sin(\zeta_i)
    \label{eq:cpg_out}
\end{align}
where $\zeta_i$ denotes the internal state, and $\theta_i$ is consistent with the reference angle of $i$-th actuator.
$\alpha$, $u_i^r$, $u_i^\eta$, and $u_i^A$ denote the internal parameters of this CPG model.
For all the CPGs (a.k.a. actuators), we set the same parameters, $\alpha = 2$, $u_i^r = 10$, $u_i^\eta = 1$, and $u_i^A = \pi / 4$, respectively.
$dt$ is the discrete time step and set to be $0.02$ sec.

Even with this CPG model, the robot has room for optimization of the stiffness of each actuator, $k_i$.
Therefore, the proposed method is applied to the optimization of $k_i \in [0, 1]$ ($i = 1, 2, \ldots, 8$).
Let us introduce the state and action spaces of the robot.

As for the state of the robot $s$, the robot observes the internal state of each actuator: $\theta_i$ angle; $\dot{\theta}_i$ angular velocity; $\tau_i$ torque; and $k_i$ stiffness (different from the command value due to control accuracy).
To evaluate its locomotion, the coordinates of its head, $x$ and $y$, are additionally observed (see Fig.~\ref{fig:snake_env}).
In addition, as mentioned before, the action of the robot $a$ is set to be $k_i$.
In summary, 34-dimensional $s$ and 8-dimensional $a$ are summarized as follows:
\begin{align}
    s &= [\theta_1, \dot{\theta}_1, \tau_1, k_1; \theta_2, \dot{\theta}_2, \tau_2, k_2; \ldots; \theta_8, \dot{\theta}_8, \tau_8, k_8; x, y]^\top
    \label{eq:robot_state} \\
    a &= [k_1, k_2, \ldots, k_8]^\top
    \label{eq:robot_action}
\end{align}

For the definition of task, i.e. the design of reward function, we consider forward locomotion.
Since the primitive motion is already generated by the CPG model, this task can be accomplished only by restraining the sideward deviation.
Therefore, we define the reward function as follows:
\begin{align}
    r(s, a) = - |y|
    \label{eq:robot_reward}
\end{align}

The proposed method learns the composed policy for the above task.
At the beginning of each episode, the robot is initialized to the same place with $\theta_i = 0$ and $k_i = 0.5$.
Afterwards, the robot starts to move forward, and if it goes outside of observable area (including a goal) or spends 2000 time steps, that episode is terminated.
We tried 100 episodes in total.

\subsubsection{Learning results}

We depict the learning curves about the score (a.k.a. the sum of rewards) and the mixture ratio in Fig.~\ref{fig:exp_result}.
Note that the moving average with 5 window size is applied to make it easier to see the learning trends.
From the score, we say that the proposed method improved straightness of the snaking locomotion.
Indeed, Fig.~\ref{fig:snap_learn}, which illustrates the snapshots of experiment before and after learning, clearly indicates that the robot could succeeded in forward locomotion only after learning.

As well as the successful trials in Fig.~\ref{fig:sim_result}, this experiment also increased the mixture ratio at first, and afterwards, the FF policy was optimized, reducing the mixture ratio toward 0.5 (but converged on around 0.7).
We found the additional feature that during 10-30 episodes, probably when the transfer of skill from the FB to FF policies was active, the score temporarily decreased.
This would be due to the increased frequency of use of the non-optimal FF policy, resulting in erroneous behaviors.
After that period, however, the score became stably high, and we expect that the above skill transfer was almost complete and the optimal actions could be sampled even from the FF policy.

\subsubsection{Demonstration with learned policies}

To see the accomplishment of the skill transfer, after the above learning, we apply the decomposed FB/FF policies individually into the robot.
On the top of Fig.~\ref{fig:snap_compare}, we shows the overlapped snapshots (red/blue robots correspond to the FB/FF policies, respectively).
With the FF policy, of course, errors in the initial state were gradually increased and accumulated, namely the two results can never be completely consistent.
However, the difference at the goal was only a few centimeters.
This result suggests that the skill transfer from the FB to FF policies has been achieved as expected, although there is room for further performance improvement.

Finally, we emulate a sensing failure for detecting the AR marker on the head.
When the robot is in the left side of the video frame, the detection of the AR marker is forcibly failed, and returns wrong (and constant) $x$ and $y$.
In that case, the FB policy would collapse, while the FF policy is never affected by the sensing failure.
On the bottom of Fig.~\ref{fig:snap_compare}, we shows the overlapped snapshots, where the left side with the sensing failure is shaded.
Until the robot left the left side, the locomotion obtained by the FB policy drifted in front of the video frame, and it was apparent that the robot could not recovered by the goal.

In detail, Fig.~\ref{fig:exp_test_w_failure} illustrates the stiffness during this test.
Note that the vertical axis is the unbounded version of $k_i$, and can be encoded into the original $k_i$ through sigmoid function.
As can be seen in the figure, the sensing failure absolutely affected the outputs by the FB policy, while the FF policy ignored it and outputted periodically.
Although this test is a proof-of-concept, it clearly shows the sensitivity/robustness of the FB/FF policies to sensing failures that may occur in real environment.
We then conclude that a framework that can learn both the FB/FF policies in a unified manner, such as the proposed method, is useful in practice.

\section{Conclusion}
\label{sec:conclusion}

In this paper, we derive a new optimization problem of both the FB/FF policies in a unified manner.
Its point is to consider minimization/maximization of the KL divergences between the trajectory, predicted by the composed policy and the stochastic dynamics model, and the optimal/non-optimal trajectories, inferred based on \textit{control as inference}.
With the composed policy as mixture distribution, the stochastic dynamics model that is approximated by variational method yields the soft regularization, i.e. the cross entropy between the FB/FF policies.
In addition, by designing the mixture ratio to prioritize the policy with higher confidence, we can expect that the FB policy is first optimized since its state dependency can easily be found, then its skill is transferred to the FF policy via the regularization.
Indeed, the numerical simulation and the robot experiment verified that the proposed method can stably solve the given tasks, that is, it has capability to optimize the composed policy even with the different learning law from the traditional RL.
In addition, we demonstrated that using our method, the FF policy can be appropriately optimized to generate the similar behavior to one with the FB policy.
As a proof-of-concept, we finally illustrated the robustness of the FF policy to the sensing failures when the AR marker could not be detected.

However, we also found that the FF policy (or its base RNNs) occasionally failed to be optimized due to the cause of extreme updates toward wrong direction.
To alleviate this problem, in the near future, we need to make the FF policy conservatively update, for example, using a soft regularization to its prior.
With more stable learning capability, the proposed method will be applied to various robotic tasks with potential for the sensing failures.


\begin{backmatter}

\section*{Availability of data and materials}
The data that support the findings of this study are available from the corresponding author, TK, upon reasonable request.

\section*{Competing interests}
The authors declare that they have no competing interests.

\section*{Author's contributions}
TK proposed the algorithm and wrote this manuscript.
KY developed the hardware and performed the experiments.

\section*{Funding}
This work was supported by Telecommunications Advancement Foundation Research Grant.


\bibliographystyle{bmc-mathphys} 
\bibliography{biblio}      




\clearpage
\section*{Figures}

\begin{figure}[!ht]
    \centering
    \includegraphics[keepaspectratio=true,width=0.95\linewidth]{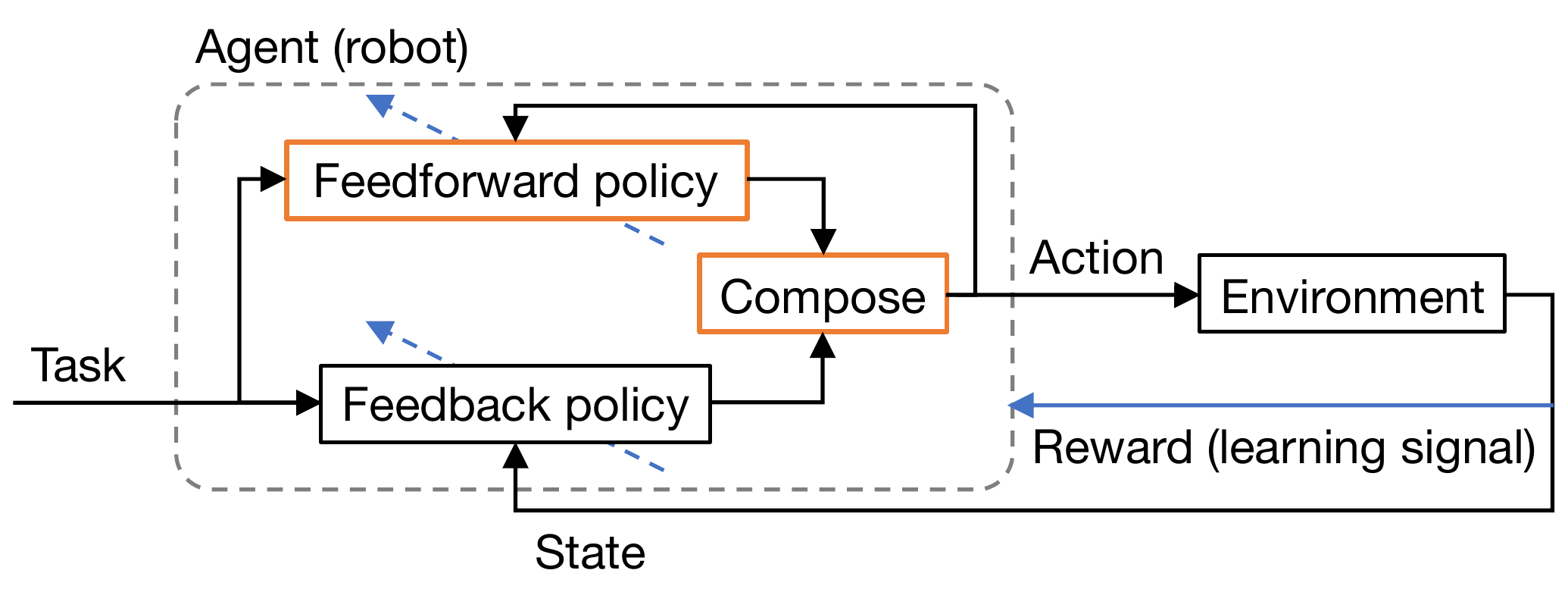}
    \caption{Proposed RL framework:
    it contains both the FB/FF policies in parallel;
    policies outputted from them are composed to sample action;
    according to reward, both the FB/FF policies are optimized in a unified manner.
    }
    \label{fig:fffb_framework}
\end{figure}

\begin{figure}[!ht]
    \centering
    \includegraphics[keepaspectratio=true,width=0.95\linewidth]{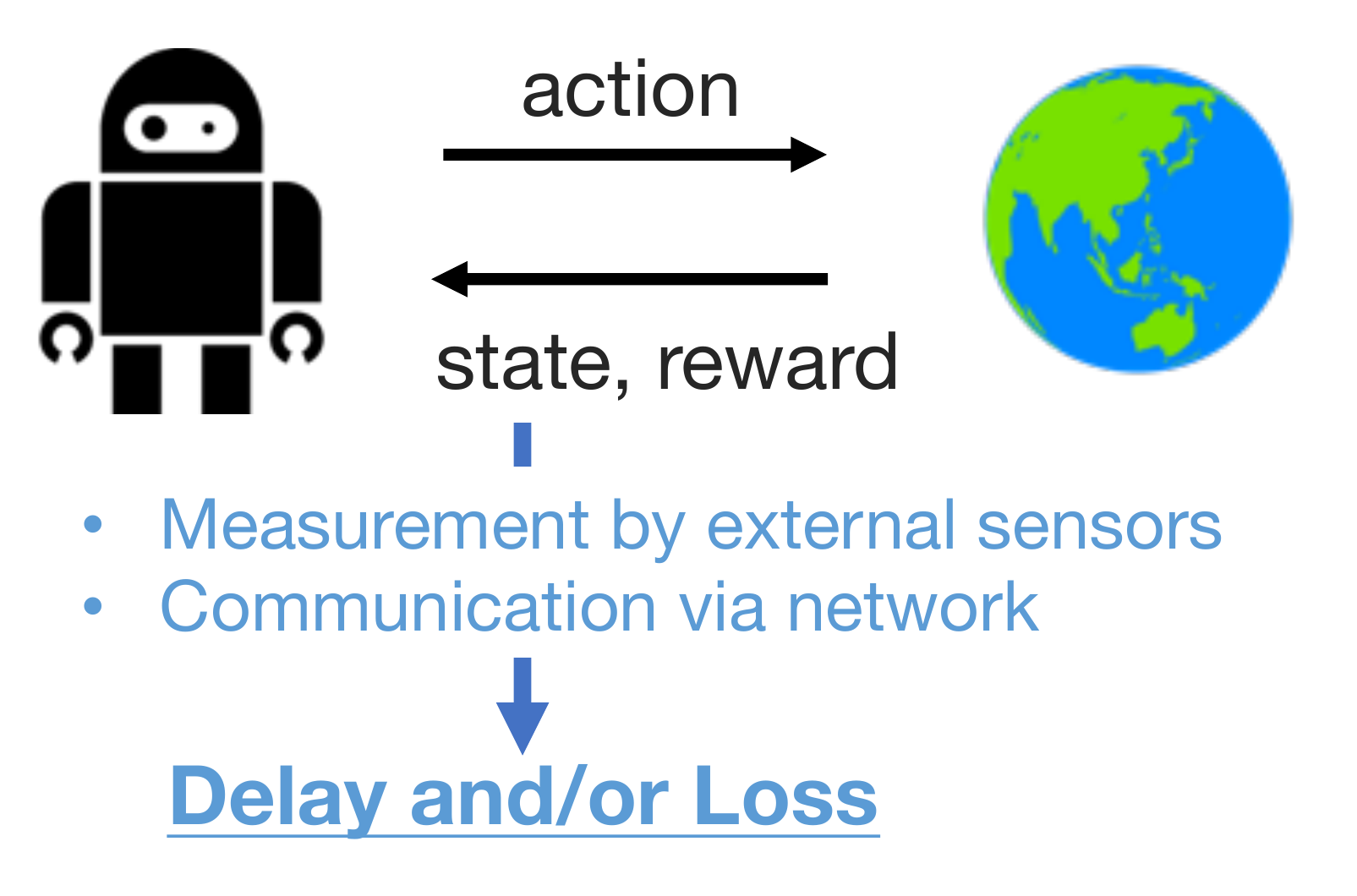}
    \caption{Loop of RL with sensing failures:
    in general RL, an agent of the left interacts with environment on the right by action sampled from policy depending on the current state;
    according to state transition probability, the new state is observed with related reward;
    however, in practice, state observation is probably with risk of sensing failures like occlusion and packet loss.
    }
    \label{fig:problem_rl}
\end{figure}

\begin{figure}[!ht]
    \centering
    \includegraphics[keepaspectratio=true,width=0.95\linewidth]{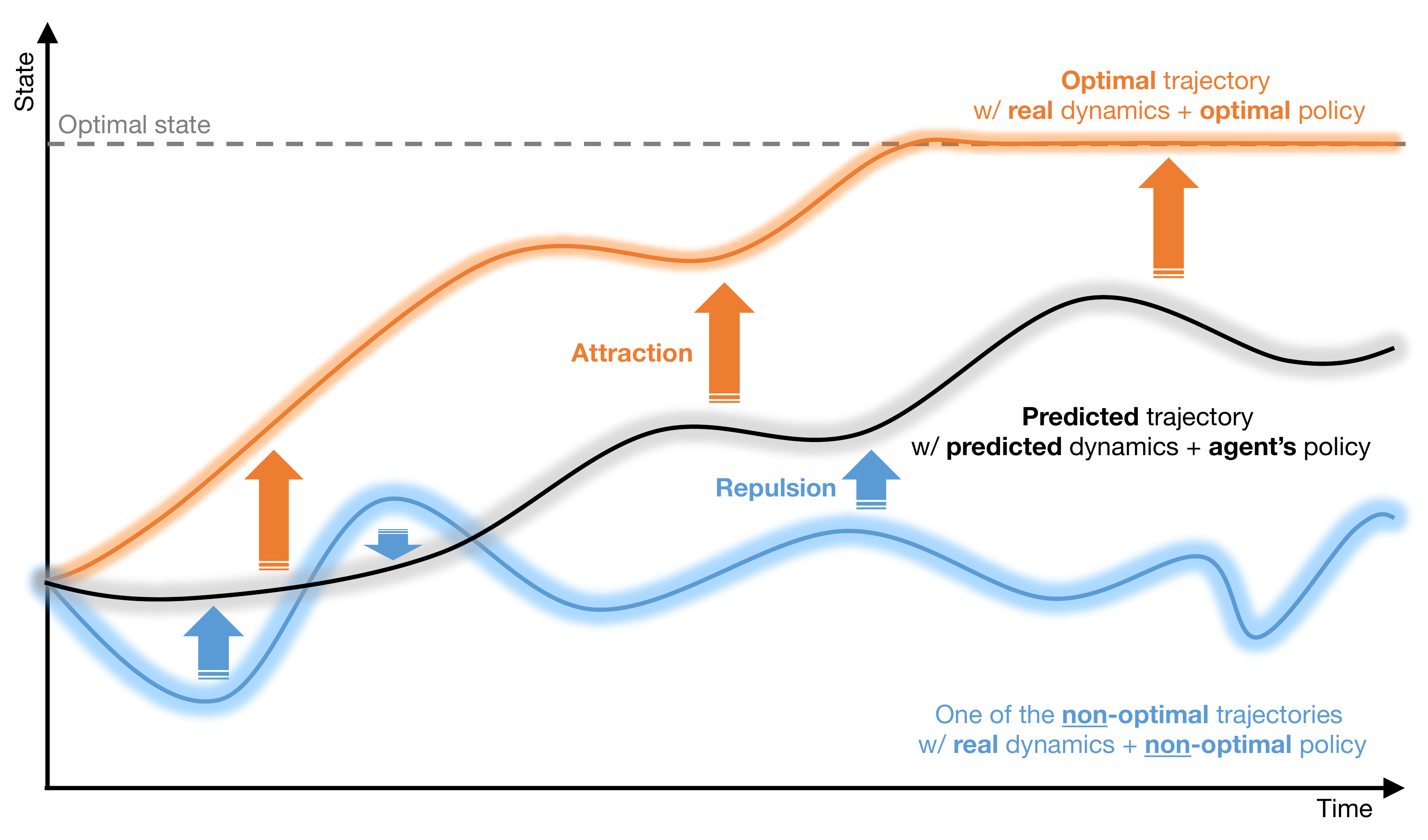}
    \caption{Trajectory optimization problem:
    the trajectory can be predicted with the composed policy and the stochastic dynamics model;
    the optimal/non-optimal trajectories can be inferred with the optimal/non-optimal policies and the true state transition probability;
    the predicted trajectory is desired to be close to the optimal trajectory, while to be away from the non-optimal trajectory;
    the divergence between trajectories can be represented by the KL divergence.
    }
    \label{fig:problem_traj_div}
\end{figure}

\begin{figure}[!ht]
    \centering
    \includegraphics[keepaspectratio=true,width=0.95\linewidth]{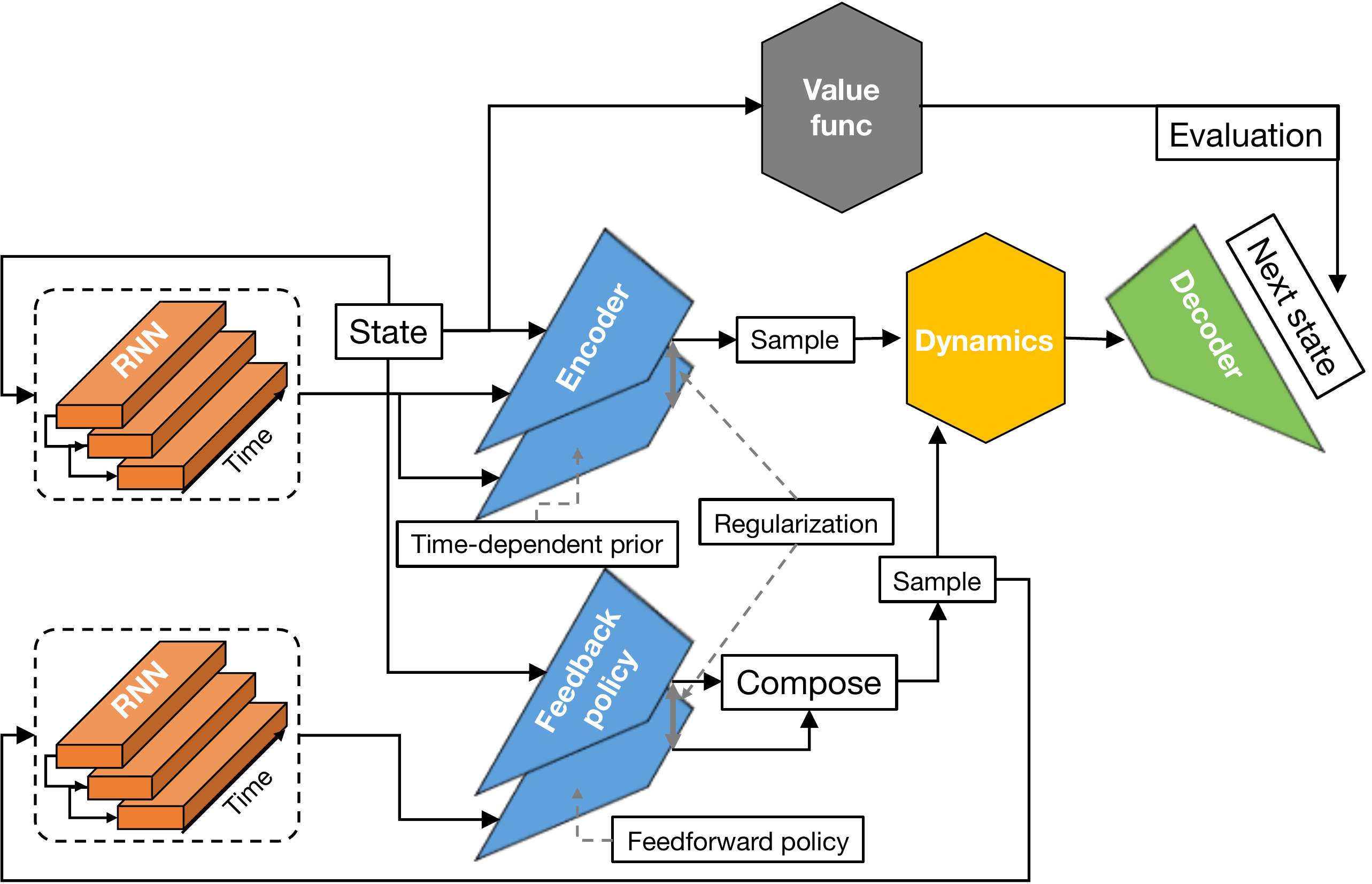}
    \caption{Network architecture of the proposed method:
    it contains seven modules for the encoder $q(z \mid s, h^s)$, decoder $p(s^\prime \mid z^\prime)$, time-dependent prior $q(z \mid h^s)$, dynamics $f(z, a)$, value function $V(s)$, and the FB/FF policies $\pi_\mathrm{FB}(a \mid s)$, $\pi_\mathrm{FF}(a \mid h^a)$ with two RNN features, $h^s$ and $h^a$;
    $\pi_\mathrm{FB}$ and $\pi_\mathrm{FF}$ are composed as $\pi$, while being regularized between each other.
    }
    \label{fig:fffb_architecture}
\end{figure}

\begin{figure}[!ht]
    \centering
    \subfigure[Score]{
        \includegraphics[keepaspectratio=true,width=0.45\linewidth]{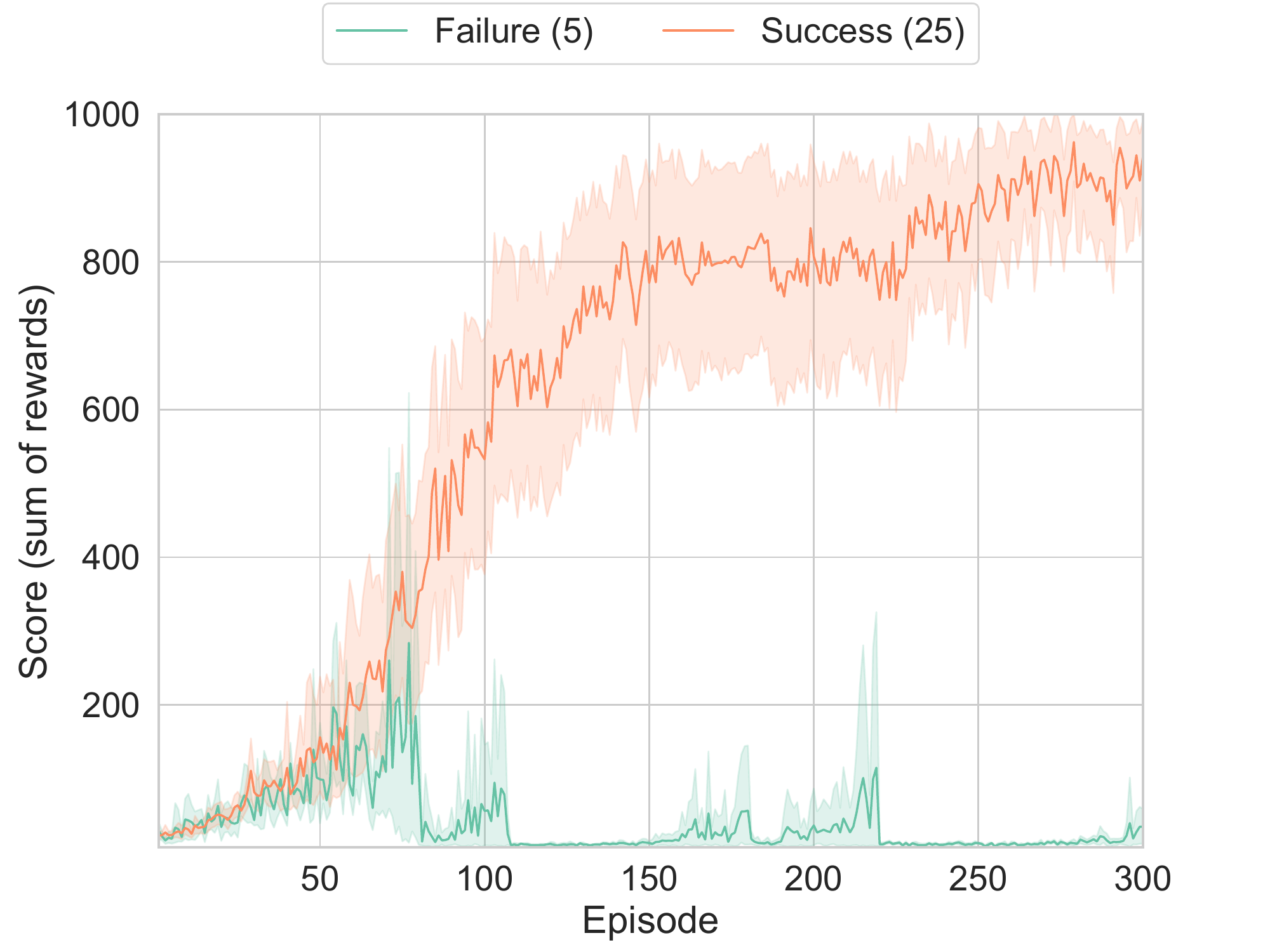}
    }
    \centering
    \subfigure[Mixture ratio]{
        \includegraphics[keepaspectratio=true,width=0.45\linewidth]{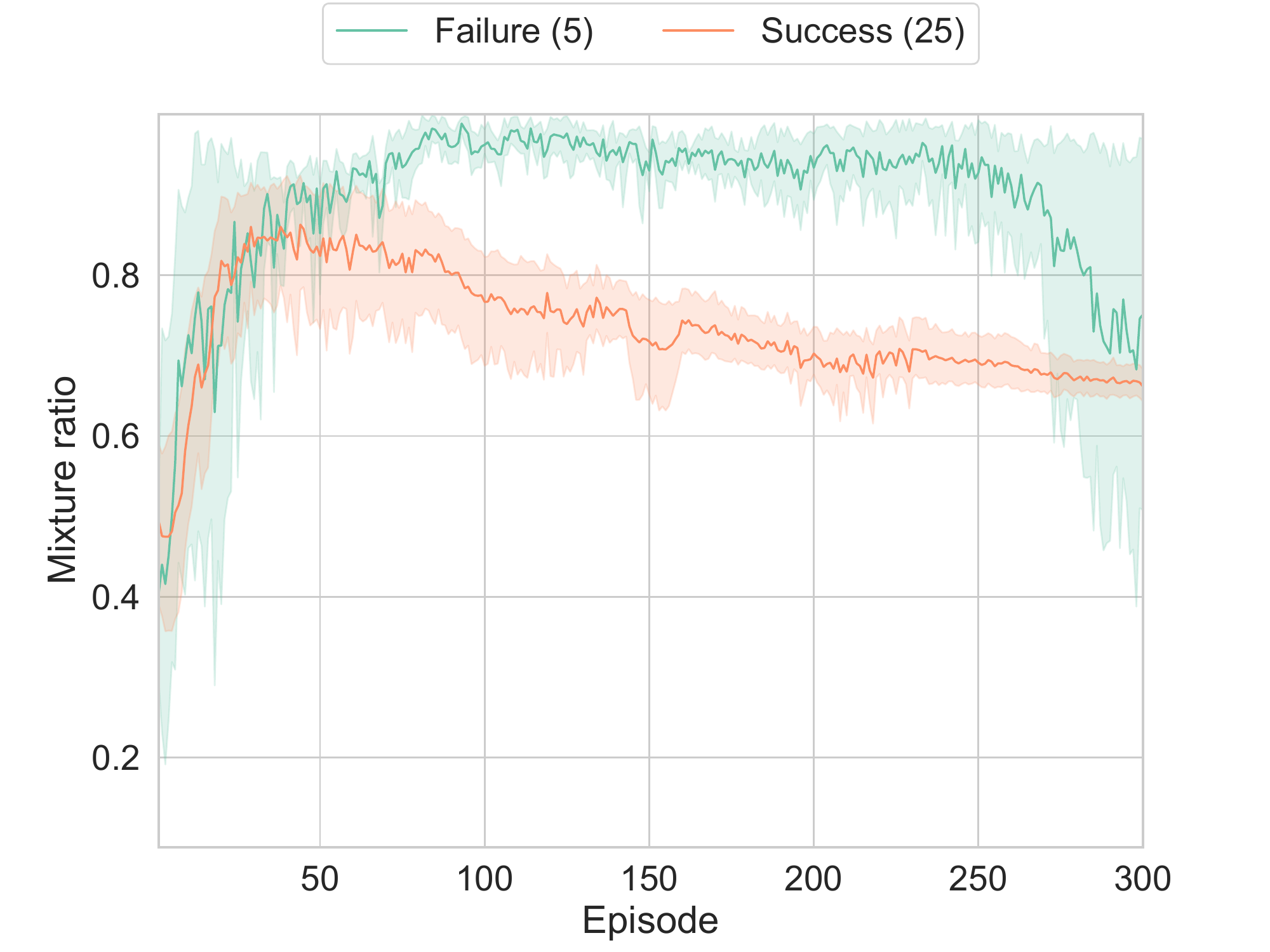}
    }
    \caption{Simulation results:
    30 trials were divided into 5 failure and 25 successful cases;
    around 150 episodes, the proposed method mostly succeeded in balancing the pole on the cart, mainly using the FB policy shown in the mixture ratio close to 1;
    afterwards, the composed policy was made deterministic with further stabilization;
    in that time, the skill of the FB policy was probably transferred into the FF policy, as can be seen in the decrease of the mixture ratio.
    }
    \label{fig:sim_result}
\end{figure}

\begin{figure}[!ht]
    \centering
    \subfigure[Distance $d$]{
        \includegraphics[keepaspectratio=true,width=0.3\linewidth]{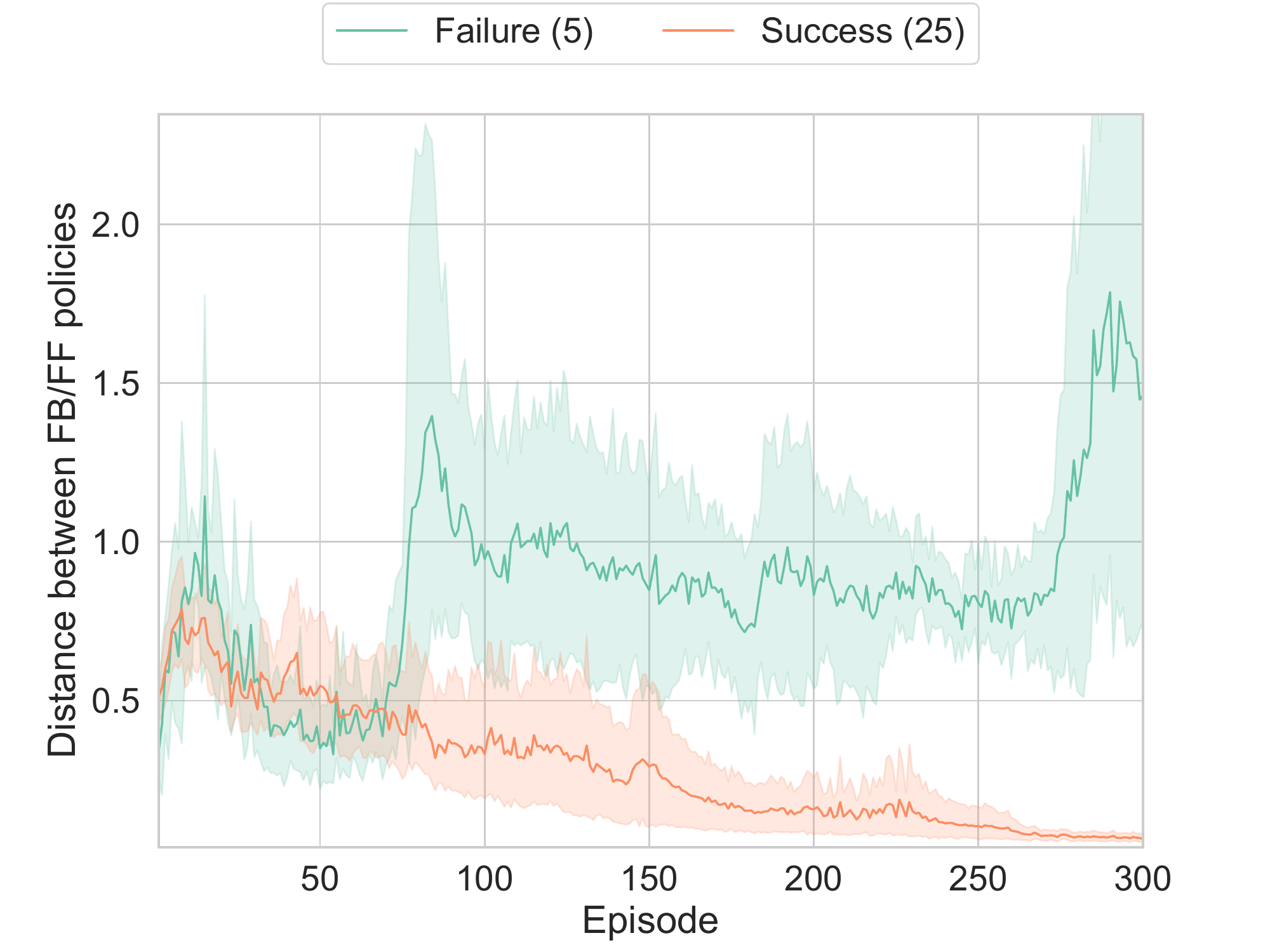}
    }
    \centering
    \subfigure[Entropy of the FB policy $H_\mathrm{FB}$]{
        \includegraphics[keepaspectratio=true,width=0.3\linewidth]{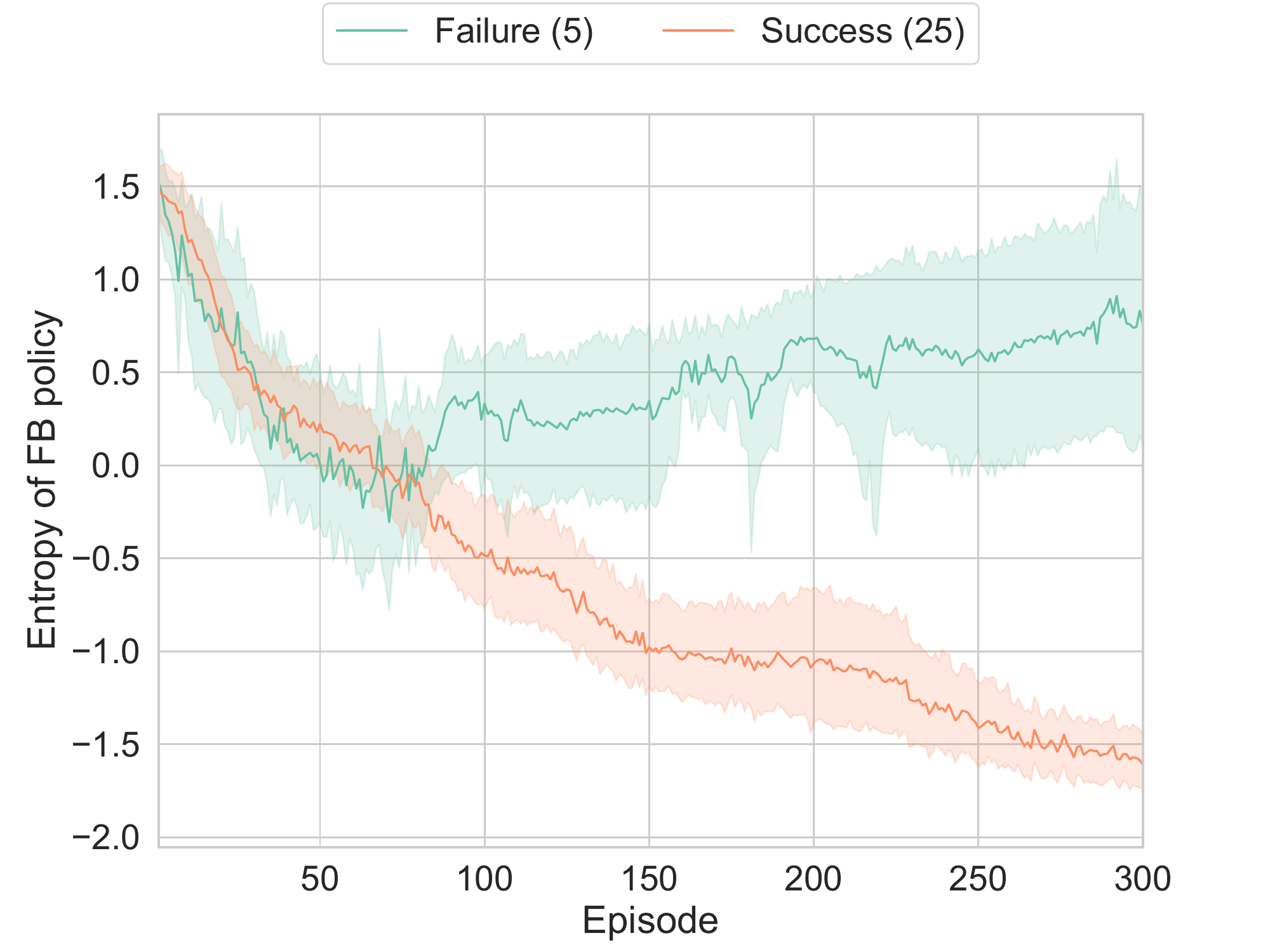}
    }
    \subfigure[Entropy of the FF policy $H_\mathrm{FF}$]{
        \includegraphics[keepaspectratio=true,width=0.3\linewidth]{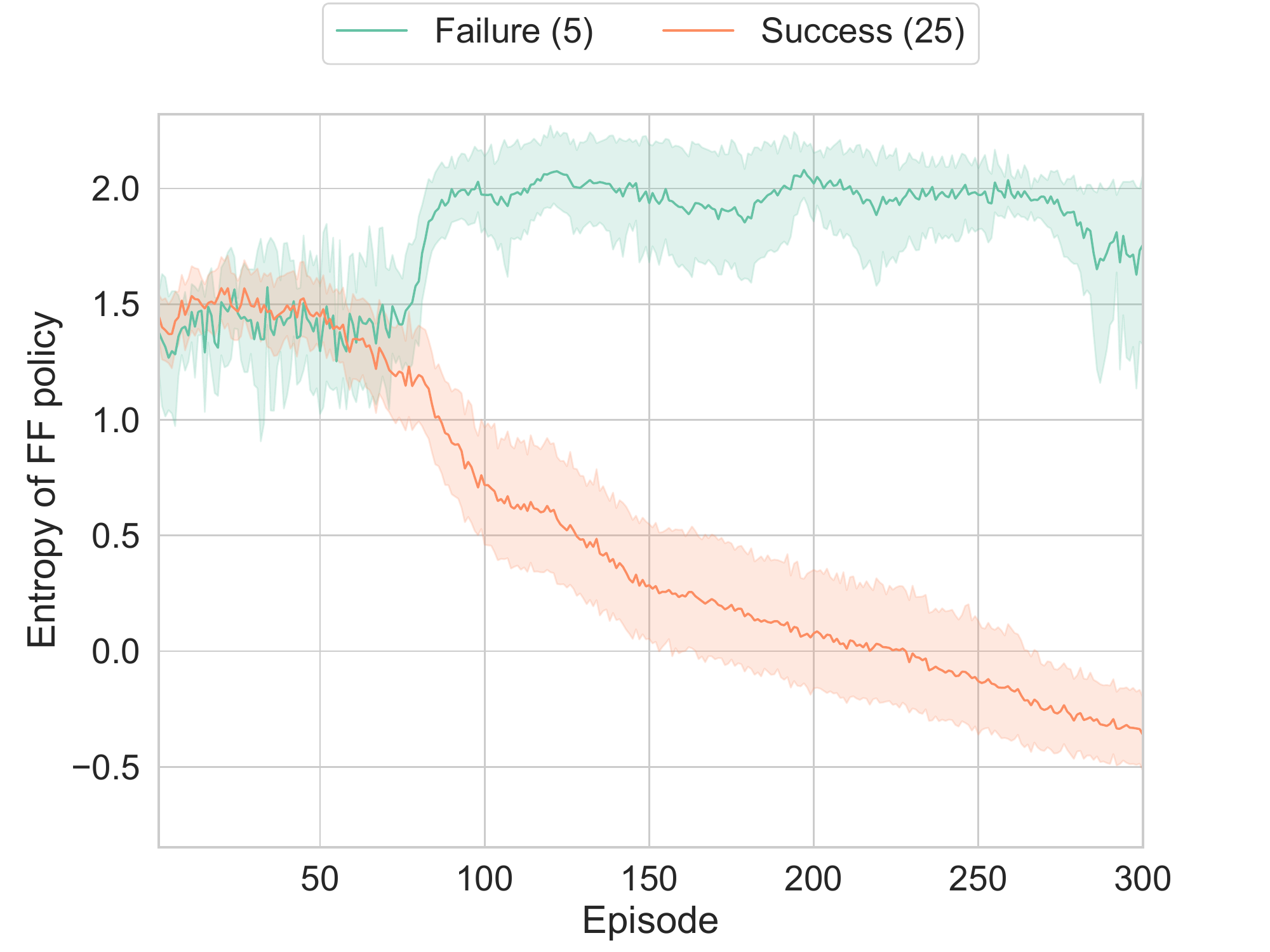}
    }
    \caption{Decomposition of mixture ratio:
    30 trials were divided into 5 failure and 25 successful cases;
    around 80th episode on the five failure cases, $d$ and $H_\mathrm{FF}$ were suddenly jumped to higher values;
    this suggests the wrong updates of the FF policy (or its base RNNs);
    according to this erroneous behavior, $H_\mathrm{FB}$ was pulled into the wrong direction by the FF policy, thereby resulting in the failures of the balancing task.
    }
    \label{fig:sim_analysis}
\end{figure}

\begin{figure}[!ht]
    \centering
    \includegraphics[keepaspectratio=true,width=0.95\linewidth]{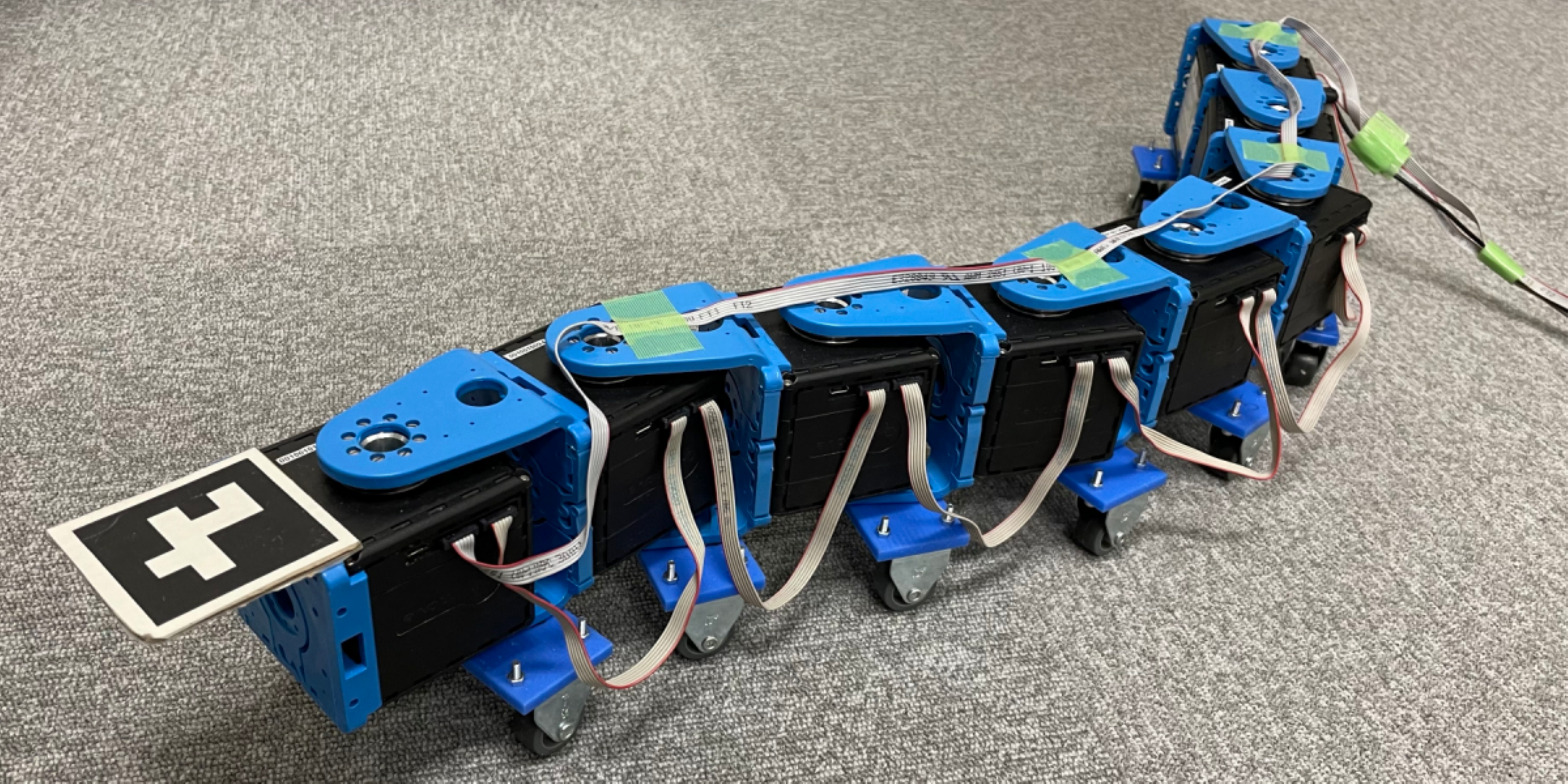}
    \caption{Snake robot with eight VSAs serially connected:
    as its actuator, we use Qbmove developed QbRobotics, which can control its stiffness;
    this robot is on casters to easily drive forward by snaking locomotion, base of which is generated by CPGs.
    }
    \label{fig:snake_robot}
\end{figure}

\begin{figure}[!ht]
    \centering
    \includegraphics[keepaspectratio=true,width=0.95\linewidth]{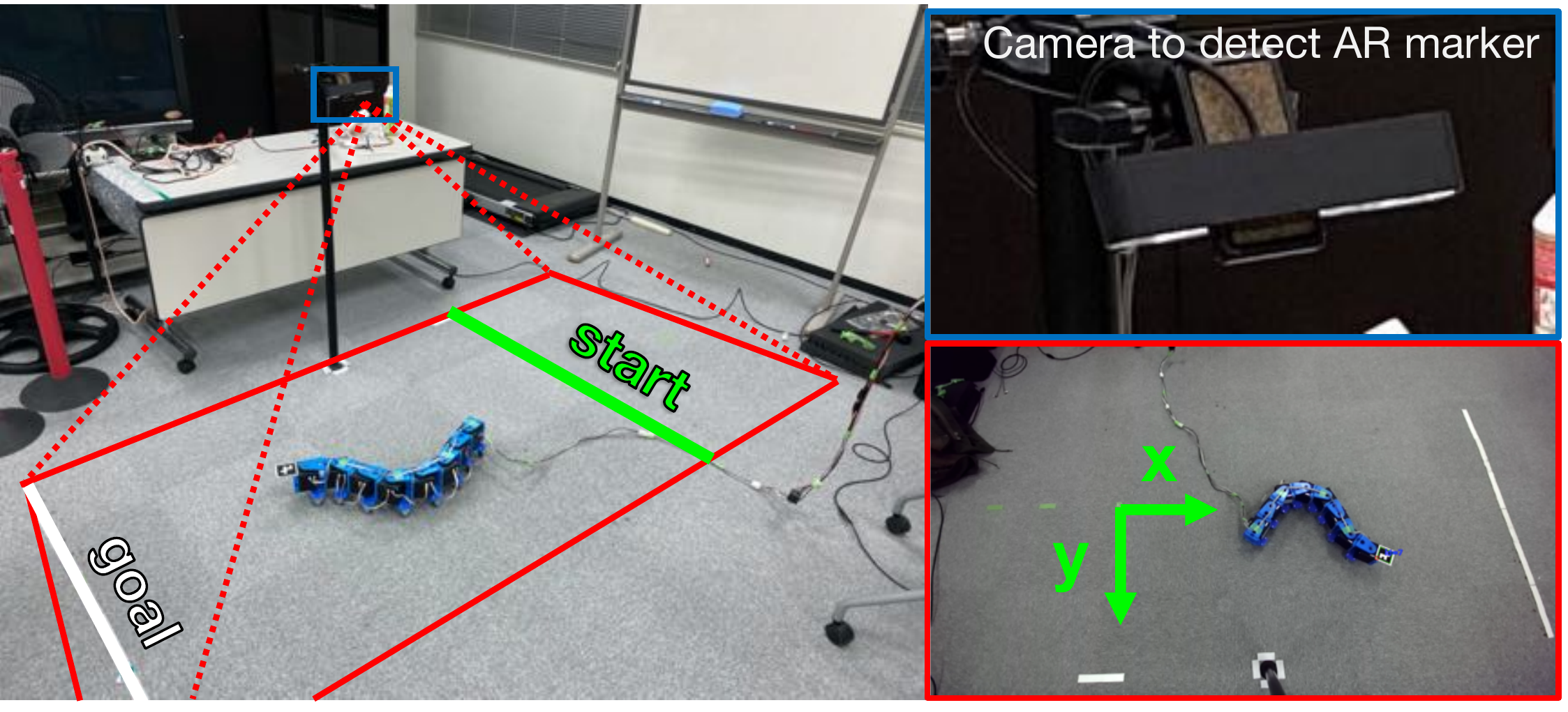}
    \caption{Experimental field:
    on the top of this field, a camera to detect the robot head by the AR marker is placed;
    by controlling the stiffness of each actuator, the robot tries to move forward, i.e. $x$-direction.
    }
    \label{fig:snake_env}
\end{figure}

\begin{figure}[!ht]
    \centering
    \subfigure[Score]{
        \includegraphics[keepaspectratio=true,width=0.45\linewidth]{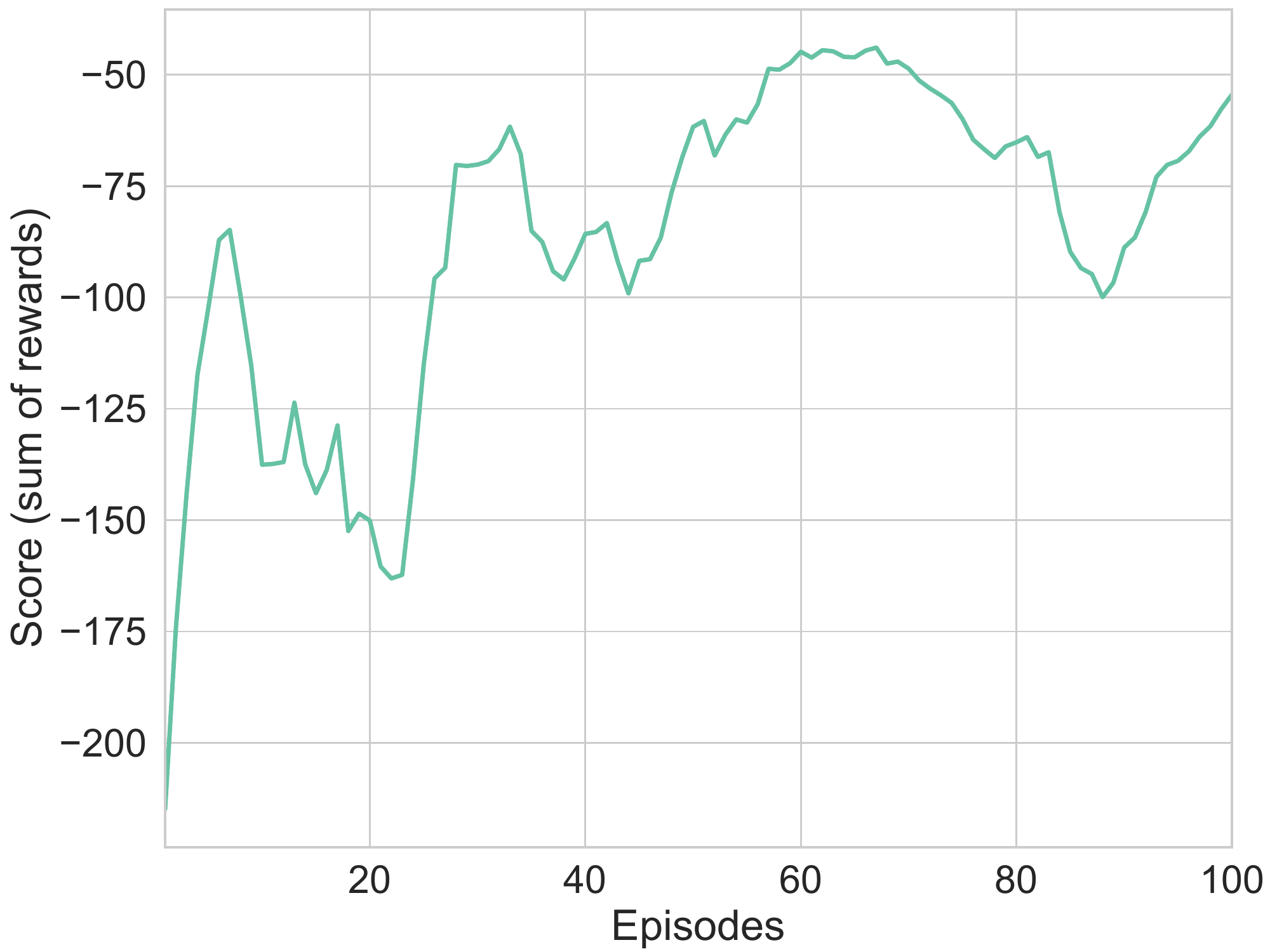}
    }
    \centering
    \subfigure[Mixture ratio]{
        \includegraphics[keepaspectratio=true,width=0.45\linewidth]{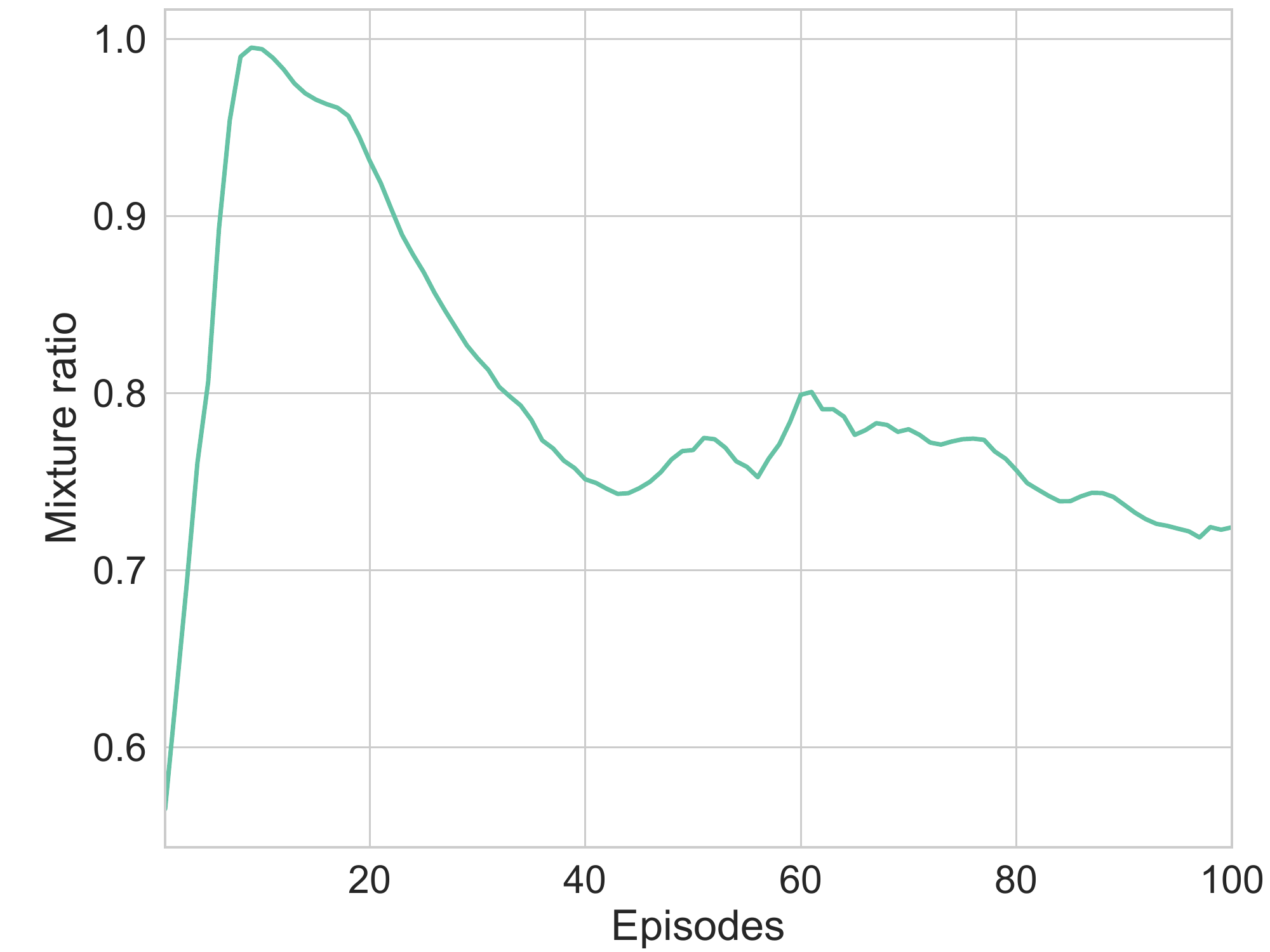}
    }
    \caption{Experimental results:
    for visibility of learning trends, moving average with 5 window size is applied;
    the proposed method successfully improved the straightness of the snaking motion by optimizing the stiffness;
    we found the skill transfer from the FB policy to the FF policy, as can be seen in the mixture ratio as well as Fig.~\ref{fig:sim_result};
    as a remarkable point, during this transfer (10-30 episodes), the score temporarily decreased probably due to the increased frequency of use of the non-optimal FF policy.
    }
    \label{fig:exp_result}
\end{figure}

\begin{figure}[!ht]
    \centering
    \includegraphics[keepaspectratio=true,width=0.95\linewidth]{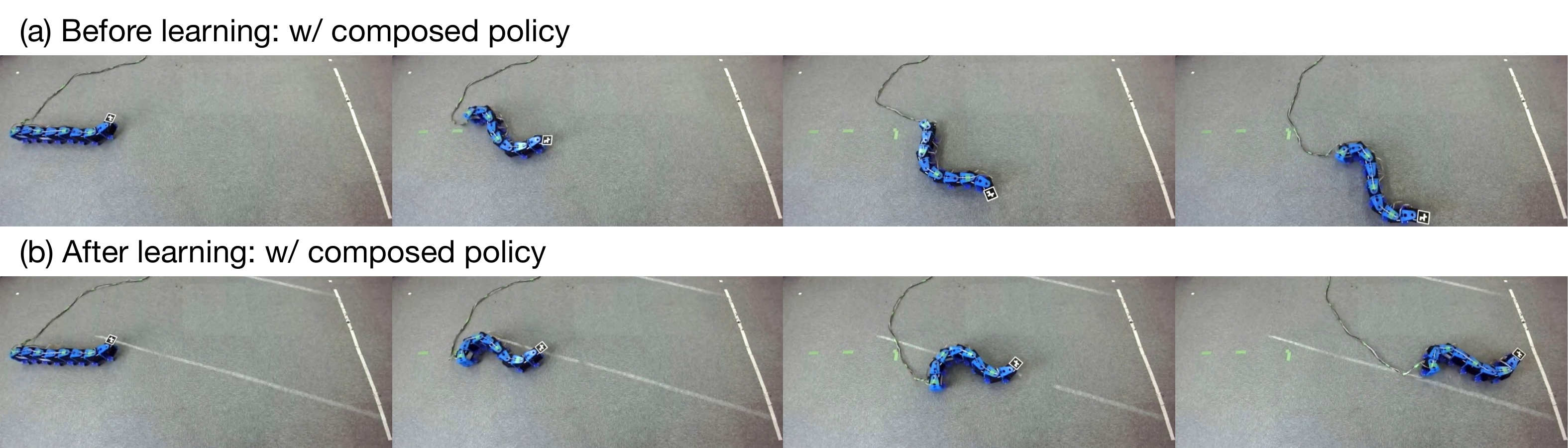}
    \caption{Snapshots before and after learning:
    before learning, the initial policy failed to make the snaking locomotion forward;
    in contrast, the proposed method yielded the forward locomotion using the optimized composed policy.
    }
    \label{fig:snap_learn}
\end{figure}

\begin{figure}[!ht]
    \centering
    \includegraphics[keepaspectratio=true,width=0.95\linewidth]{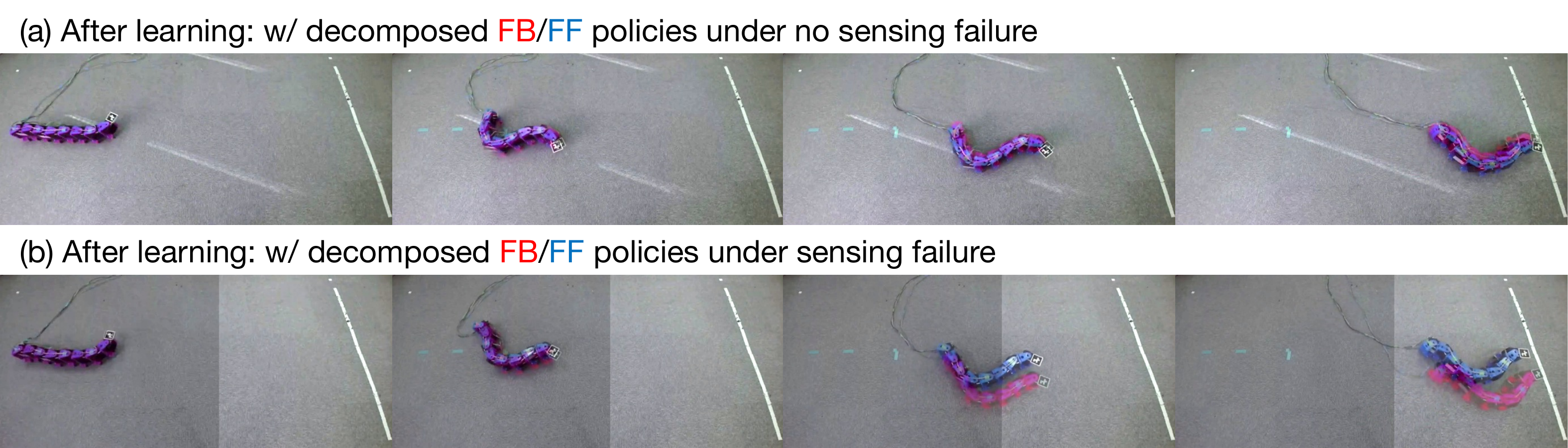}
    \caption{Snapshots with/without the sensing failures:
    the robot was controlled by the decomposed FB (red) or FF (blue) policy;
    without the sensing failures, both the policies generated almost the same forward locomotion, which indicates the proper skill transfer;
    with the sensing failures to detect the AR marker, indicated as the shaded area, the FB policy drifted the robot to the side due to the wrong signal;
    in contrast, the FF policy could achieve the forward locomotion by ignoring the wrong signal in principle.
    }
    \label{fig:snap_compare}
\end{figure}

\begin{figure}[!ht]
    \centering
    \includegraphics[keepaspectratio=true,width=0.95\linewidth]{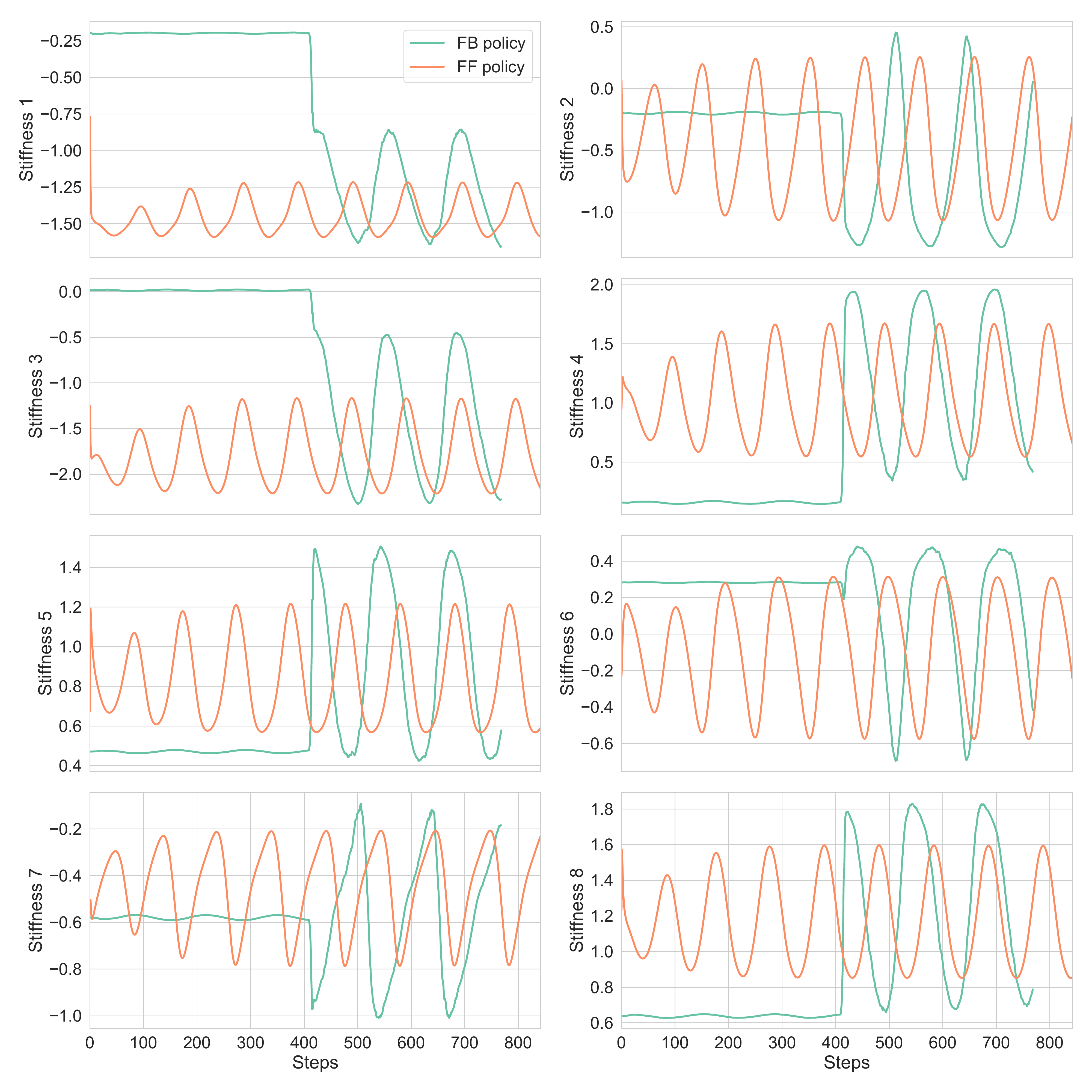}
    \caption{Stiffness of each actuator when the sensing failures were intentionally caused:
    the vertical axis depicts the unbounded version of $k_i$, which can be encoded by sigmoid function;
    during the sensing failures, the FB policy outputted obviously erroneous stiffness;
    in contrast, the FF policy could hold the periodic outputs;
    note that the phase and amplitude deviations in the area without the sensing failures can be attributed to incomplete skill transfer and recovery attempts from lateral deviation.
    }
    \label{fig:exp_test_w_failure}
\end{figure}


\clearpage
\section*{Tables}

\begin{table}[!ht]
    \caption{Parameter configuration}
    \label{tab:parameter}
    \centering
    \begin{tabular}{ccc}
        \hline\hline
        Symbol & Meaning & Value
        \\
        \hline
        $|Z|$ & Dimension size of latent space & 6
        \\
        $\beta_T$ & Inverse temperature & 10
        \\
        $\beta_z$ & Weight of regularization in $z$ & 1e-2
        \\
        $\beta_a$ & Weight of regularization in $a$ & 1e-4
        \\
        $\eta$ & Remaining computational graph & 1e-4
        \\
        $\gamma$ & Discount factor & 0.99
        \\
        $\alpha$ & Learning rate & 3e-4
        \\
        $\rho$ & Echo state property~\cite{gallicchio2018design} & 0.5
        \\
        $(\tau, \nu)$ & Hyperparameters for t-soft update~\cite{kobayashi2021t} & (0.5, 4.0)
        \\
        $(\lambda_\mathrm{max}^1, \lambda_\mathrm{max}^2, \kappa)$ & Hyperaparameters for adaptive eligibility traces~\cite{kobayashi2020adaptive} & (0.5, 0.95, 10)
        \\
        \hline\hline
    \end{tabular}
\end{table}

\clearpage
\section*{Additional Files}
\subsection*{Additional file 1 --- Experimental video}
This video summarized all the experiments using the snake robot for forward snaking locomotion.
At first, we confirmed that the constant (maximum, more specifically) stiffness failed the forward locomotion to clarify the necessity of its optimization.
At the beginning of learning, the robot could not keep the forward locomotion naturally.
By learning with the proposed method, the robot could achieve the forward locomotion by using the composed policy.
Even with the decomposed FB (red) or FF (blue) policy, we found almost the same motion.
However, when the detection failure was intentionally applied, the FB policy failed to keep the locomotion forward, while the FF policy could do so.

\end{backmatter}

\end{document}